\documentclass[letterpaper]{article} 
\usepackage{aaai2027}  
\nocopyright
\usepackage[hyphens]{url}  
\usepackage{graphicx} 
\usepackage{natbib}  
\usepackage{caption} 
\usepackage{amsmath,amssymb,amsfonts}
\usepackage{mathtools}
\usepackage{booktabs}
\usepackage{array}
\usepackage{tabularx}
\usepackage{listings}
\usepackage{multirow}
\usepackage{tikz}
\usetikzlibrary{arrows.meta,calc}
\definecolor{lotCodeKeyword}{RGB}{21, 89, 156}
\definecolor{lotCodeString}{RGB}{151, 84, 17}
\definecolor{lotCodeComment}{RGB}{80, 118, 72}
\definecolor{lotCodeBuiltin}{RGB}{128, 64, 140}
\definecolor{lotCodeNumber}{RGB}{112, 112, 112}
\definecolor{lotCodeRule}{RGB}{188, 188, 188}
\definecolor{lotInk}{RGB}{35, 42, 52}
\definecolor{lotMuted}{RGB}{104, 116, 132}
\definecolor{lotPanel}{RGB}{248, 250, 252}
\definecolor{lotAxis}{RGB}{79, 91, 107}
\definecolor{lotDense}{RGB}{38, 92, 166}
\definecolor{lotSparse}{RGB}{214, 99, 48}
\definecolor{lotBP}{RGB}{38, 145, 108}
\definecolor{lotBand}{RGB}{229, 239, 255}
\definecolor{lotPolicy}{RGB}{255, 236, 211}
\definecolor{lotWarn}{RGB}{190, 70, 62}

\lstdefinestyle{pythonlisting}{
  language=Python,
  basicstyle=\ttfamily\scriptsize,
  keywordstyle=\color{lotCodeKeyword},
  stringstyle=\color{lotCodeString},
  commentstyle=\color{lotCodeComment},
  columns=fullflexible,
  keepspaces=true,
  breaklines=true,
  breakatwhitespace=false,
  showstringspaces=false,
  tabsize=4,
  numbers=left,
  stepnumber=1,
  numberstyle=\tiny\ttfamily\color{lotCodeNumber},
  numbersep=6pt,
  numberblanklines=true,
  xleftmargin=2.3em,
  framexleftmargin=1.9em,
  frame=tb,
  framerule=0.35pt,
  framesep=3pt,
  rulecolor=\color{lotCodeRule},
  aboveskip=0.75\baselineskip,
  belowskip=0.75\baselineskip,
  captionpos=t
}

\graphicspath{{Figures/}}

\newcommand{\E}{\mathbb{E}}
\newcommand{\Acc}{\mathrm{Acc}}
\newcommand{\ECE}{\mathrm{ECE}}
\newcommand{\NLL}{\mathrm{NLL}}
\newcommand{\AUROC}{\mathrm{AUROC}}
\newcommand{\AUC}{\mathrm{AUC}}
\newcommand{\IDFAR}{\mathrm{IDFAR95}_{\mathrm{OOD}}}
\newcommand{\MIA}{\mathrm{MIA}_{\mathrm{ex}}}
\newcommand{\RepDist}{\mathrm{RepDist}}
\newcommand{\RobAcc}{\mathrm{RobAcc}}
\newcommand{\DBE}{D_{\mathrm{BE}}}
\newcommand{\sel}{\mathrm{sel}}
\newcommand{\DBEacc}{\DBE^{-\mathrm{acc}}}
\newcommand{\SelDBE}{\widehat{D}_{\mathrm{BE}}^{-\mathrm{acc},\sel}}
\newcommand{\SelAcc}{\widehat{\Acc}^{\sel}_{\mathrm{clean}}}

\newtheorem{theorem}{Theorem}
\newtheorem{proposition}{Proposition}

\title{Lottery Tickets Are Not Deployment Tickets}
\author{Bum Jun Kim\thanks{Corresponding author}}
\affiliations{The University of Tokyo\\
bumjun.kim@weblab.t.u-tokyo.ac.jp}

\begin{document}

\maketitle

\begin{abstract}
	Reports on how sparsification, compression, and lottery tickets change model behavior have been mixed in the prior literature, with beneficial effects observed in some studies and adverse effects in others. Moreover, prior work has not considered actual deployment conditions, where decision logic is already fixed for the incumbent. To assess these mixed findings from a practical standpoint, we study the production-replacement question at the deployment level, namely whether an accuracy-matched lottery ticket or another sparse challenger can replace an incumbent dense model without reconfiguring downstream decision logic. We therefore audit a broad, protocol-specific panel of deployment-relevant behaviors spanning calibration, OOD response, class-level reliability, representations, and downstream policy decisions, and summarize clean-accuracy-excluded deviations with a behavioral-compatibility distance. Across experiments on CIFAR-10, CIFAR-100, Imagenette, Flowers-102, and FGVC-Aircraft with ResNet, WideResNet, ConvNeXt, and Vision Transformer backbones, sparse candidates repeatedly recover dense-reference accuracy yet remain behaviorally different; in several study-band-matched settings, LTs also show lower corruption accuracy. In small-gap settings with fixed-threshold policy diagnostics, lottery-ticket replacement changes 7\% to 10\% of accept--review decisions. This churn creates precisely the burden that drop-in replacement is meant to avoid: reconfiguring and revalidating downstream decision logic. These findings establish the limits of clean-accuracy certification: Establishing compatibility with a fixed incumbent is distinct from attributing churn uniquely to sparsity or treating every measured deviation as harmful. Our theory explains the routing result: Even exact pointwise top-1 agreement cannot bound fixed-threshold decision changes, and small confidence shifts near the operating boundary can generate first-order routing churn. Clean-accuracy recovery therefore does not certify policy-compatible sparse replacement; the accuracy-recovery check must be followed by an operating-point compatibility audit and signed service-level checks.
\end{abstract}

\section{Introduction}
\citet{frankle2019lottery} introduced the lottery ticket (LT) hypothesis, which states that dense neural networks often contain sparse subnetworks that can be trained in isolation to recover comparable test accuracy. Subsequent work connected the success of late-rewound tickets to early-training stability \citep{frankle2020linear}, compared weight and learning-rate rewinding with fine-tuning \citep{renda2020comparing}, and established large-scale sparsity benchmarks \citep{gale2019state}. Sparse solutions also transfer across datasets and downstream tasks \citep{morcos2019one,iofinova2022sparse}. The promise extends beyond clean accuracy: Pruning can act as a regularizer that improves generalization \citep{jin2022pruning}, compression can improve out-of-distribution (OOD) robustness \citep{diffenderfer2021winning}, sparsity can preserve adversarial robustness and shrink the robust generalization gap \citep{sehwag2020hydra,chen2022sparsity}, and calibration-aware recipes improve ticket reliability \citep{venkatesh2020calibrate,lei2023calibrating}. Taken together, these studies advocate LTs and other sparse models as practically promising replacements.

\begin{figure*}[t!]
	\centering
	\includegraphics[width=\textwidth]{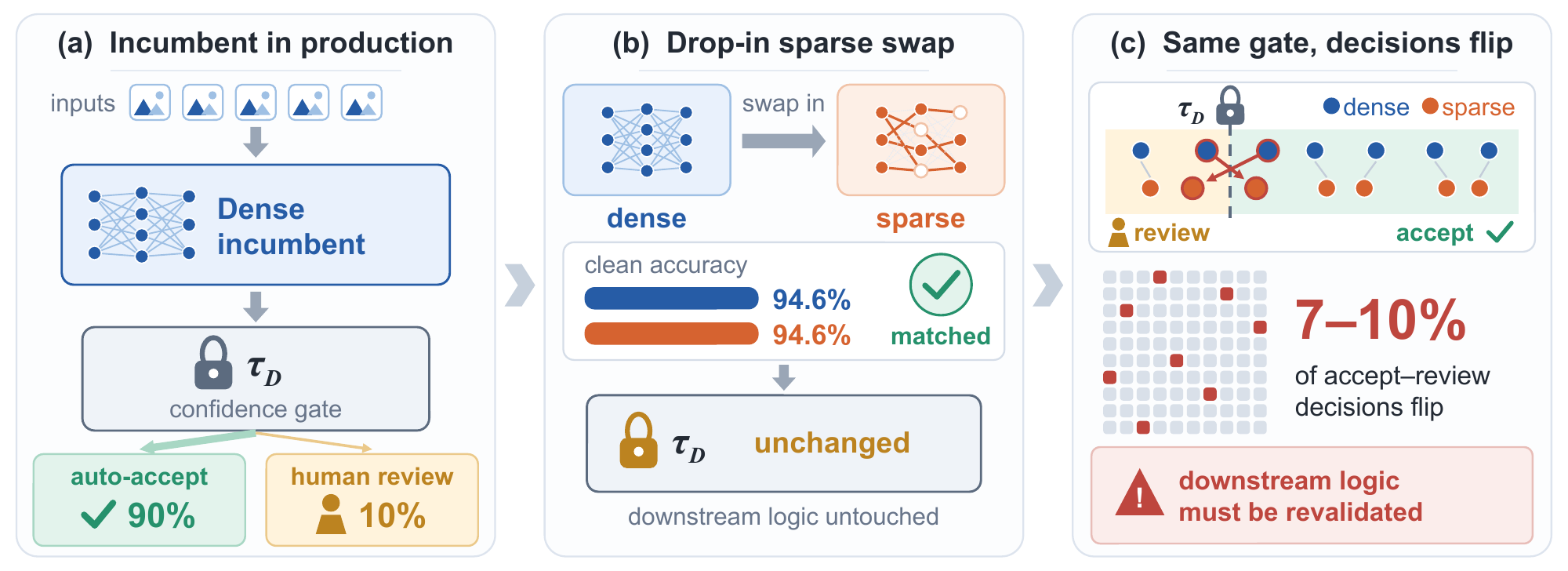}
	\caption{The production-replacement scenario at the operating-point level. The diagram first shows a deployment that configures a confidence gate for the dense incumbent. Inputs whose confidence falls below the cutoff $\tau_D$ are routed to human review under a fixed review budget, and the rest are auto-accepted by the downstream logic. Next, a sparse challenger that has recovered the incumbent's clean accuracy is swapped in as a strict drop-in replacement, leaving $\tau_D$ and the inherited downstream logic unchanged. Finally, the same frozen gate still produces different actions because confidence can cross $\tau_D$ even when the top-1 label does not. Crossings occur in both directions, so the review rate can be preserved while individual routing decisions change. $\mathrm{AcceptFlip}_{\tau_D}$ reaches 7\% to 10\% in small-gap settings with fixed-threshold policy diagnostics.}
	\label{fig:overview}
\end{figure*}

Other studies, however, report that sparsification changes model behavior: Aggregate accuracy can conceal changes in per-example errors, group-level effects, calibration, robustness, and uncertainty under shift \citep{hooker2020compressed,liebenwein2021lost,tran2022disparate, dutta2024accuracy,chen2022wineverything,tong2026compression}. Most directly, \citet{chen2022wineverything} find that, at appropriate sparsity levels, accuracy-preserving LTs can match or outperform their dense counterparts in distribution-shift generalization, uncertainty, interpretability, and loss geometry, although generalization to certain shifts and uncertainty are more sensitive to sparsification. Model-update work likewise shows that aggregate improvement can coexist with backward-incompatible errors or user-facing churn \citep{bansal2019updates,srivastava2020backward}. Therefore, evidence about the behavior and practical standing of sparse models is mixed, favorable in some studies and adverse in others.

Beyond these mixed reports, prior work also offers little validation under the conditions that an actual deployment imposes. Such a deployment fixes a confidence threshold and a review budget for the incumbent and commits downstream components to the resulting accept--review actions. Existing sparse evaluations hold neither the threshold nor the review budget fixed, leaving a gap that clean-accuracy recovery cannot close. \citet{chen2022wineverything} recompute each diagnostic separately for the sparse subnetwork, so the evaluation grades an LT as a freshly commissioned model rather than as a replacement inserted into decision logic already fixed for the incumbent; an LT can therefore pass every such diagnostic and still change what that logic does. \citet{mitra2024investigating} likewise report calibration and corruption robustness of post-hoc pruned networks as standalone benchmark metrics, but an aggregate calibration score does not reveal whether confidence mass has moved across the particular cutoff a deployment consumes, so unchanged or improved calibration is compatible with substantially changed routing. \citet{tong2026compression} restore coverage with a conformal benchmark by calibrating each compressed model. This model-specific recalibration may be appropriate, but the recalibration constitutes a coordinated model-and-policy update rather than a strict drop-in swap: The new rule can require renewed validation of review capacity, accepted risk, monitoring, and application-specific approval. These procedures consume personnel time and operational resources, imposing costs beyond the model swap itself. Our audit asks whether an accuracy-matched challenger can be deployed free from that work. \citet{chee2022model} and \citet{dutta2024accuracy} do compare a compressed model against the corresponding baseline example by example, yet the compared object is the model output rather than the action that inherited downstream logic takes on that output. Neither agreement constraints enforced during compression nor post-compression measurements of answer flips or output distances constrain the routing decision, because an unchanged answer can still cross a fixed confidence gate while a changed answer that both models send to review costs the deployment nothing.

Deployment supplies exactly such logic: Production classifiers feed confidence-based selective policies \citep{geifman2017selective}, OOD rejection and shift monitoring \citep{hendrycks2017baseline}, robustness checks \citep{hendrycks2019benchmarking}, and monitoring and retraining pipelines \citep{sculley2015hidden}. Such gates are used in practice: Amazon Rekognition triggers human review for content moderation \citep{amazon2026a2i}, and a classifier deployed in the Very Long Baseline Array Fast Radio Transients Experiment reserved low-confidence transient candidates for review \citep{wagstaff2016machine}. Because models with similar held-out accuracy can still differ in calibrated confidence, rejected inputs, class-specific errors, and internal representations \citep{breiman2001statistical,damour2022underspecification}, clean-accuracy recovery leaves these inherited dependencies untested. Figure~\ref{fig:overview} summarizes the resulting replacement question.

In summary, prior work not only offers mixed evidence on the benefits and drawbacks of LTs and other sparse models but also leaves a deployment-validation gap: Existing evaluations do not test sparse challengers against the incumbent operating point for which a running system is already configured. To address this gap, we study whether accuracy-matched LTs and other sparse models can serve as strict drop-in replacements and introduce a method-agnostic operating-point compatibility audit. The challenger must first satisfy a clean-accuracy constraint; the audit then separates two release questions: signed service-level preservation and backward-compatible continuity with the incumbent. The audit pairs each challenger with the corresponding dense reference, holds the downstream decision rule fixed, measures accept--review changes under a dense-derived threshold, and summarizes the remaining active coordinates with a clean-accuracy-excluded behavioral-compatibility distance.

Our central result is that clean-accuracy recovery does not certify compatibility with an incumbent operating point. Empirically, accuracy-matched sparse candidates remain measurably different from the corresponding dense references on active audit coordinates, and in several study-band-matched settings LTs also have lower corruption accuracy. The theory shows that even exact pointwise top-1 agreement cannot bound fixed-threshold decision changes and characterizes how confidence movement near the threshold produces routing flips. Together, the audit, theory, and experiments shift the replacement question from clean-accuracy recovery alone to preservation of the incumbent operating point and relevant signed service levels.

\section{Behavioral Compatibility Audit}

This section makes the strict drop-in replacement criterion operational in two steps. We first describe each sparse candidate using a clean-accuracy-excluded behavior vector that covers calibration, OOD response, dense-threshold policy decisions, and combined reference and reliability diagnostics. We then impose clean accuracy as a separate matching constraint and measure the standardized distance between the candidate's behavior and that of the dense reference. The distance supplies a comparative compatibility screen, while signed coordinates supply service-level evidence. Statistical equivalence and application-specific utility are evaluated against separate, predeclared release criteria.

Throughout, policy denotes the downstream rule that maps model confidence to an operational action. The dense reference is the matched, trained, unpruned model, and a sparse candidate is a trained sparse alternative evaluated as a potential replacement. We use network and subnetwork when the architectural or connectivity distinction is relevant. We collectively call the inherited downstream dependencies and their acceptable tolerances the replacement contract. The contract's contents are application specific.

\subsection{Behavioral Coordinates}

\paragraph{Coordinate-Selection Principle.}
We choose coordinates based on the downstream dependency that each coordinate probes, not because every coordinate is required in every deployment. The calibration block audits confidence consumers; the OOD and corruption coordinates audit distribution-shift handling; the policy block audits an inherited accept--review gate; and the reference and reliability block audits class-specific service levels, privacy exposure, per-example output continuity, and, when intermediate features are consumed, representation compatibility. Each quantity can change while clean aggregate accuracy remains fixed. The vector therefore defines a protocol-specific audit panel for common replacement dependencies, with application-specific coordinates and costs supplied by the replacement contract. This audit panel contains two logically distinct types of evidence. Signed service-level coordinates, such as corruption accuracy and worst-class accuracy, support directional service-level comparisons. Reference-paired compatibility coordinates, such as prediction or routing disagreement, measure continuity with the incumbent.

\paragraph{Clean-Accuracy-Excluded Behavior Vector.}
Let $(X,Y)$ denote a clean input--label pair with class set $\mathcal{Y}\coloneqq\{1,\ldots,K\}$, let $f_D$ be a dense reference classifier, and let $f$ be a sparse candidate classifier. For any predictor $g$, let $p_g(y\mid x)$ be the predictive probability assigned by $g$, $\hat y_g(x)\coloneqq\arg\max_y p_g(y\mid x)$ the prediction made by $g$ under a fixed tie-breaking rule, and $c_g(x)\coloneqq\max_y p_g(y\mid x)$ the maximum confidence produced by $g$. Let $\tau_D$ be the dense-threshold policy cutoff defined below. Clean aggregate accuracy, $\Acc(g)\coloneqq\Pr[\hat y_g(X)=Y]$, is evaluated separately as an explicit matching constraint. Throughout, clean-accuracy-excluded means excluding only this matched clean aggregate accuracy; accuracy-like diagnostics such as worst-class accuracy and corruption accuracy ($\RobAcc$) remain behavioral coordinates. The coordinate set below includes expected calibration error (ECE), negative log-likelihood (NLL), OOD area under the receiver operating characteristic curve (AUROC), in-distribution false-alarm rate at 95\% OOD recall ($\IDFAR$), membership inference attack (MIA) area under the curve (AUC) excess, and $\RobAcc$. Detailed definitions follow.

The clean-accuracy-excluded behavior vector is written as a concatenation of calibration, OOD, dense-threshold policy, and combined reference and reliability coordinates:
\begin{align*}
	\mathbf{b}^{-\mathrm{acc}}(f;f_D,\tau_D)
	 & \coloneqq\operatorname{concat}(
	\mathbf{b}_{\mathrm{cal}},
	\mathbf{b}_{\mathrm{OOD}},
	\mathbf{b}_{\mathrm{pol}},
	\mathbf{b}_{\mathrm{ref}}),
\end{align*}
where repeated coordinate arguments are suppressed and
\begin{align*}
	\mathbf{b}_{\mathrm{cal}}
	 & \coloneqq(\ECE,\NLL,\mathrm{Brier}),                              \\
	\mathbf{b}_{\mathrm{OOD}}
	 & \coloneqq(\AUROC_{\mathrm{OOD}},\IDFAR,
	\mathrm{Conf}_{\mathrm{OOD}}),                                       \\
	\mathbf{b}_{\mathrm{pol}}
	 & \coloneqq(\mathrm{ReviewRate}_{\tau_D},\mathrm{AutoErr}_{\tau_D}, \\
	 & \mathrm{OODAccept}_{\tau_D},\mathrm{AcceptFlip}_{\tau_D}),        \\
	\mathbf{b}_{\mathrm{ref}}
	 & \coloneqq(\mathrm{WorstClassAcc},\MIA,\RepDist,                   \\
	 & \mathrm{PredDisagree},\RobAcc).
\end{align*}
The complete specification includes coordinate-by-coordinate motivation and exact definitions, together with the distance-fit, normalization, active-coordinate, missing-value, and scale-fallback rules used below. In particular, $\mathrm{PredDisagree}$ measures top-1 disagreement with the dense reference, and $\RobAcc$ measures corruption accuracy on a corruption distribution.

\paragraph{Incumbent-Threshold Policy Coordinates.}
We also compute decision-level diagnostics when the required validation and test logits and OOD evaluation data are available. We treat a confidence gate configured for the dense reference as part of the inherited replacement contract. We therefore freeze the dense-derived threshold rather than recalibrating the threshold for each candidate: Candidate-specific recalibration would define a coordinated model-and-policy update rather than the strict drop-in replacement audited here. In our audit, the fixed binary gate auto-accepts an input when the evaluated model's confidence on that input is at or above $\tau_D$ and otherwise routes the input to review; policy compatibility means preserving the incumbent's per-input accept--review actions after model replacement under this unchanged gate. Because inputs below the threshold are sent to review, the dense reference sets $\tau_D$ to the empirical 10\% quantile of the dense reference's validation confidences, $\tau_D\coloneqq q_{0.10}(c_{f_D}(X_{\mathrm{val}}))$, so that approximately the lowest-confidence 10\% are reviewed. Finite-sample interpolation and ties can make the realized review rate differ from this nominal 10\% target. Production policies may use different thresholds and review budgets; the 10\% target supplies a common, auditable, and reproducible proxy operating point. A single prespecified operating point is sufficient for our certification question: If clean-accuracy recovery alone were a sufficient drop-in certificate, an accuracy-matched candidate would need to preserve the inherited action at the audited point. Threshold sweeps extend this test to policy robustness across operating points. Applying the same fixed threshold to a candidate gives $\mathrm{ReviewRate}_{\tau_D}$, $\mathrm{AutoErr}_{\tau_D}$, $\mathrm{OODAccept}_{\tau_D}$, and $\mathrm{AcceptFlip}_{\tau_D}$. With $(X,Y)$ drawn from the in-distribution test distribution and $Z$ drawn from the OOD distribution, define $a_f(x)\coloneqq\mathbf{1}\{c_f(x)\geq\tau_D\}$ and $a_D(x)\coloneqq\mathbf{1}\{c_{f_D}(x)\geq\tau_D\}$, and let $A_f\coloneqq\{c_f(X)\geq\tau_D\}$. Then
\begin{align*}
	\mathrm{ReviewRate}_{\tau_D}(f)
	 & \coloneqq\Pr[c_f(X)<\tau_D],             \\
	\mathrm{AutoErr}_{\tau_D}(f)
	 & \coloneqq\Pr[\hat y_f(X)\neq Y\mid A_f], \\
	\mathrm{OODAccept}_{\tau_D}(f)
	 & \coloneqq\Pr[c_f(Z)\geq\tau_D],          \\
	\mathrm{AcceptFlip}_{\tau_D}(f,f_D)
	 & \coloneqq\Pr[a_f(X)\neq a_D(X)].
\end{align*}
The complete policy specification also gives the directional decomposition, missing-value convention, and operational interpretation of these coordinates. $\mathrm{AcceptFlip}_{\tau_D}$ is the fraction of in-distribution examples whose acceptance or review decision changes relative to the dense reference.

\subsection{Behavioral-Compatibility Distance}

\paragraph{Diagonal Behavioral-Compatibility Distance.}
Within each protocol group $G$, the active behavioral coordinates are standardized using fixed group-specific normalization statistics to obtain $Z_j(f)$. Let $\mathcal{M}_G$ denote the resulting active coordinate set. When $|\mathcal{M}_G|>0$, the diagonal behavioral-compatibility distance is
\begin{align*}
	\DBEacc(f,f_D)
	 & \coloneqq
	\left(
	\frac{1}{|\mathcal{M}_G|}\sum_{j\in\mathcal{M}_G}
	[
		Z_j(f)-Z_j(f_D)
	]^2
	\right)^{1/2}.
\end{align*}
If $|\mathcal{M}_G|=0$, the distance is undefined and is reported as missing. Applying the same direction multiplier to both the candidate and the dense reference causes the multiplier to cancel from the compatibility distance. The direction is used only in separate utility-signed summaries. Thus, zero distance means equality of the protocol's imputed active-coordinate vectors. When the same construction is computed from held-out validation estimates of the coordinates used for selection, we denote the resulting validation distance by $\SelDBE(f,f_D)$. The Appendix specifies the missing-value rule and reports rank- and covariance-aware sensitivity variants of this summary.

\paragraph{Compatibility Rather Than Utility.}
The behavioral-compatibility distance quantifies the magnitude of the measured behavioral difference between a candidate and the deployed dense reference. A coordinate can improve and still indicate changed replacement behavior. Consequently, a nonzero distance alone does not establish worse deployment performance. Directional degradation claims in this paper instead rely on signed coordinates, most directly when a candidate's $\RobAcc$ is lower than that of the corresponding dense reference. We therefore report $\DBEacc$ together with signed coordinate values in evaluation results, whereas validation-time selection uses $\SelDBE$. Hereafter, behavioral distance is shorthand for $\DBEacc$ or for $\SelDBE$ when validation data are explicitly discussed.

\section{Accuracy Cannot Certify Policy Compatibility}

A sparse candidate can match the dense model's clean accuracy but still be incompatible with the incumbent policy. This section establishes why: Top-1 outputs do not identify confidence-based behavior, and threshold policies can amplify confidence movement near the policy threshold. Theorem~\ref{thm:policy-nonidentifiability} formalizes the resulting non-identifiability even under exact pointwise top-1 agreement. Proposition~\ref{prop:threshold-flip-bound} and Theorem~\ref{thm:threshold-amplification} then characterize finite and local threshold-flip behavior, respectively.

\begin{theorem}[Dense-threshold policy non-identifiability]
	\label{thm:policy-nonidentifiability}
	Let $f_D$ be a $K$-class predictor, $K\geq2$, with measurable predictive probabilities and a measurable tie-broken prediction $\hat y_{f_D}$, and fix $\tau_D\in(1/K,1)$. Let $(X,Y)$ have a joint clean law with a nonatomic input marginal $P_X$, and define $A_D\coloneqq\{x:c_{f_D}(x)\geq\tau_D\}$. For every $\alpha\in[0,1]$, there exists a predictive distribution $f_\alpha$ such that
	\begin{align*}
		\hat y_{f_\alpha}(x)                       & =\hat y_{f_D}(x)\text{ for every }x, \\
		\Acc(f_\alpha)                             & =\Acc(f_D),                          \\
		\mathrm{PredDisagree}(f_\alpha,f_D)        & =0,                                  \\
		\mathrm{AcceptFlip}_{\tau_D}(f_\alpha,f_D) & =\alpha.
	\end{align*}
\end{theorem}

This result goes beyond the familiar observation that equal aggregate accuracy can hide different errors: Even exact pointwise top-1 agreement places no nontrivial universal bound on changes to the inherited dense-threshold policy.

Nonatomicity serves only to permit arbitrary real-valued rates; under the uniform empirical law on $n$ distinct, independently addressable examples, the same construction realizes every multiple of $1/n$. This result is therefore stronger than the replacement condition used in our experiments. The theorem establishes an information limit of the usual LT certificate rather than a typical-case prediction for every LT: The certificate contains no information that rules out such a worst case. In particular, the observed acceptance-flip rates between 7\% and 10\% are compatible even with zero prediction disagreement: The label can remain fixed while confidence crosses $\tau_D$.

Theorem~\ref{thm:policy-nonidentifiability} is a deliberately distribution-free worst-case result. We next quantify threshold flips, first with a finite-perturbation bound and then with the exact local coefficient linking confidence displacement along a specified local predictor path to policy flips. Define the dense threshold margin and the candidate confidence displacement by
\begin{align*}
	D_\tau   & \coloneqq c_{f_D}(X)-\tau_D, &
	\Delta_c & \coloneqq c_f(X)-c_{f_D}(X).
\end{align*}
Unless stated otherwise, probabilities and expectations in the next two results are under the clean-input law $P_X$.

\begin{proposition}[Finite threshold-flip bound]
	\label{prop:threshold-flip-bound}
	For every pair $(f,f_D)$ and every $\varepsilon>0$,
	\begin{align}
		\mathrm{AcceptFlip}_{\tau_D}(f,f_D)
		 & =\Pr[D_\tau\geq0>D_\tau+\Delta_c]\nonumber  \\
		 & +\Pr[D_\tau<0\leq D_\tau+\Delta_c]\nonumber \\
		 & \leq
		\Pr[|D_\tau|\leq\varepsilon]
		+\Pr[|\Delta_c|>\varepsilon].
		\label{eq:threshold-flip-sandwich}
	\end{align}
	Consequently, for every $p>0$ with $\E|\Delta_c|^p<\infty$, the final term admits the moment bound
	\begin{align*}
		\mathrm{AcceptFlip}_{\tau_D}(f,f_D)
		 & \leq \Pr[|D_\tau|\leq\varepsilon]
		+\frac{\E|\Delta_c|^p}{\varepsilon^p}.
	\end{align*}
\end{proposition}

For the local analysis, let $K\geq2$, let $f_D$ be a $K$-class predictor, and suppose $\tau_D\in(1/K,1)$. Consider any local path $\{f_t:t>0\}$ of predictive distributions with a measurable pre-clipping confidence direction $V=V(X)$ such that
\begin{align*}
	c_{f_t}(X)=\Pi_{[1/K,1]}(c_{f_D}(X)+tV),
\end{align*}
where $\Pi_{[1/K,1]}$ denotes clipping to the feasible maximum-confidence range. Because $\tau_D$ is interior to that range, clipping does not alter the acceptance decision. For every $z\in\mathbb{R}$,
\begin{align*}
	\Pi_{[1/K,1]}(z)\geq\tau_D
	\Longleftrightarrow z\geq\tau_D.
\end{align*}

\begin{theorem}[Threshold-boundary amplification]
	\label{thm:threshold-amplification}
	For the local predictor path above, suppose $(D_\tau,V)$ has a joint density with a version $h(d,v)$ such that, for almost every $v$, $d\mapsto h(d,v)$ is continuous at zero. Assume $V$ is essentially bounded and that there exist $d_0>0$ and a measurable $\bar h:\mathbb{R}\to[0,\infty)$ such that $\int_{\mathbb{R}}|v|\bar h(v)\mathrm{d}v<\infty$ and, for almost every $v$, $h(d,v)\leq\bar h(v)$ for every $|d|\leq d_0$. Then
	\begin{align}
		\lim_{t\to0^+}
		\frac{\mathrm{AcceptFlip}_{\tau_D}(f_t,f_D)}{t}
		 & =\kappa_{\tau_D}
		\coloneqq\int_{\mathbb{R}}|v|h(0,v)\mathrm{d}v.
		\label{eq:threshold-first-order}
	\end{align}
\end{theorem}

Equivalently, Eq.~\ref{eq:threshold-first-order} gives the right-sided expansion
\begin{align*}
	\mathrm{AcceptFlip}_{\tau_D}(f_t,f_D)
	=t\kappa_{\tau_D}+o(t),
\end{align*}
which has a nonzero first-order term exactly when $\kappa_{\tau_D}>0$. We call $\kappa_{\tau_D}$ the threshold-amplification coefficient.

The relevant boundary is the confidence level set rather than the top-1 class boundary. These results motivate the reported $\mathrm{AcceptFlip}_{\tau_D}$ coordinate and the behavior-aware selection specified in the experimental setup below. The Appendix also gives an extended operational interpretation of all three results.

\section{Experiments}

\subsection{Experimental Setup}

We conduct the primary evaluation on the 10-class Canadian Institute for Advanced Research dataset (CIFAR-10) with an 18-layer residual network (ResNet-18) \citep{krizhevsky2009learning,he2016deep} at 50\%, 80\%, 90\%, and 95\% sparsity. Replications cover the 100-class variant, CIFAR-100, with ResNet-18, and CIFAR-10 with ResNet-34 and a 28-layer wide residual network (WideResNet-28-2) \citep{zagoruyko2016wide}. For pretrained evaluations, we use Imagenette \citep{deng2009imagenet} with ConvNeXt-Tiny \citep{liu2022convnet} and the Tiny variant of the Vision Transformer (ViT), denoted ViT-Tiny \citep{dosovitskiy2021image}, followed by Flowers-102 \citep{nilsback2008automated} and Fine-Grained Visual Classification of Aircraft (FGVC-Aircraft) \citep{maji2013fine} transfer evaluations.

We evaluate accuracy-only selection and the Behavior-Preserving Lottery Ticket (BP-LT) procedure over each eligible sparse pool. The pools and reported baselines include LT, LT reinitialization, which tables abbreviate as LT reinit, global and layerwise random sparse controls, and one-shot and staged magnitude pruning. We abbreviate magnitude pruning as MP and use One-shot MP and Staged MP in tables. Protocol-specific pools additionally include iterative magnitude pruning (IMP), Single-Shot Network Pruning (SNIP), and the Rigged Lottery (RigL).

\paragraph{LT Construction.}
For the reported LT baseline, we first train the matched dense reference to completion, rank all maskable weights globally by their final absolute values, and retain the exact top-$k$ entries required by the target sparsity. This ranking procedure produces a one-shot global magnitude mask. We then instantiate the sparse candidate at the protocol-specific rewinding point and train the surviving weights with the mask enforced throughout optimization. In from-scratch protocols, the rewinding point is the original initialization of the matched dense reference; in pretrained protocols, the rewinding point is the pretrained backbone together with the classifier head's original random initialization. LT reinitialization uses the same mask but a separately sampled initialization. Dense seed, mask seed, candidate training seed, and candidate identifier are stored separately in the result metadata.

For sparsity $s$, let $\mathcal{F}_s\coloneqq\{f_i\}_{i=1}^n$ denote the fixed eligible pool for one matched dense-reference seed, sparsity, and protocol cell, from which BP-LT selects one existing model. All candidates are selected using held-out validation metrics, and final metrics are recomputed on the test split. Write $\Delta_i^{\sel} \coloneqq|\SelAcc(f_i)-\SelAcc(f_D)|/\epsilon_{\mathrm{acc}}$ for the normalized absolute validation accuracy gap, where $\SelAcc$ denotes validation clean accuracy and $\epsilon_{\mathrm{acc}}>0$ is the accuracy-band tolerance, so the validation accuracy band comprises the candidates with $\Delta_i^{\sel}\leq1$.

\paragraph{BP-LT.}
BP-LT selects the in-band candidate minimizing validation behavioral-compatibility distance plus a small normalized accuracy-gap penalty with weight $\lambda$: When the validation accuracy band is nonempty, the selected candidate is
\begin{align*}
	f_{\mathrm{BP}}
	 & \in
	\operatorname*{argmin}_{\substack{f_i\in\mathcal{F}_s: \\
				\Delta_i^{\sel}\leq1}}
		[
			\SelDBE(f_i,f_D)
			+
			\lambda
			(\Delta_i^{\sel})^2
		].
\end{align*}
Inside the band, behavioral closeness is the primary term, while the quadratic accuracy penalty discourages choosing a candidate near the edge of the allowed band. The penalty is zero at an exact validation-accuracy match and equals $\lambda$ at the band boundary. This penalty structure implements a soft trade-off within a hard eligibility screen. If the band is empty, the fallback reverses the emphasis by making the normalized accuracy gap the primary term and squared behavioral distance the secondary term. The fallback uses deterministic tie-breaking and excludes nonfinite validation values. For all reported selections, we set $\epsilon_{\mathrm{acc}}=0.015$ and $\lambda=0.2$.

We stratify protocol--method groups by recovery quality: A group is study-band-matched when the group's dense reference passes the dataset-specific accuracy floor and every evaluated test accuracy gap is finite and lies within the prespecified test recovery band of $\pm1.5$ percentage points (pp), near-recovery when the group fails that rule with an absolute mean gap of at most 3~pp, and recovery-stress otherwise. Raw gaps remain reported throughout, so narrower application-specific margins can be applied. The standard CIFAR protocols use Street View House Numbers (SVHN) \citep{netzer2011reading} for OOD evaluation and the CIFAR-10 corruption benchmark (CIFAR-10-C) when available.

The Appendix provides the full experimental protocol and reproducibility details, additional results and analyses, complete audit-coordinate and distance specifications, proofs, BP-LT eligibility, fallback, and implementation details, and further discussion, including limitations.

\subsection{CIFAR-10 ResNet-18 Results}

\paragraph{Multi-Sparsity Evidence.}

Table~\ref{tab:cifar10-sparsity} reports the primary CIFAR-10 ResNet-18 multi-sparsity evaluation. The dense reference reaches $(94.62\pm0.26)\%$ accuracy. At 50\% sparsity, the LT baseline matches the dense reference's mean accuracy within 0.01~pp yet has a mean behavioral distance of $\DBEacc=0.648$ from that reference. On the protocol-standardized scale, this distance is a root-mean-square separation of about 0.65 units across the active behavioral coordinates. BP-LT keeps the mean accuracy gap near zero and reduces the distance to $0.446$. At 80\% sparsity, the LT baseline remains close in accuracy and has a distance of $0.501$. Accuracy-only selection is much farther away with a distance of $0.890$. BP-LT reduces the distance to $0.437$ with a 0.35~pp mean accuracy gap. At 90\% and 95\%, all sparse methods become more strained, but BP-LT still has lower behavioral distance than the LT baseline and accuracy-only selection on average. The signed $\RobAcc$ result at 80\% is stronger than symmetric incompatibility alone: Although the LT is only 0.16~pp below the dense reference in clean accuracy, the LT's $\RobAcc$ falls from $65.71\%$ to $64.14\%$, a 1.57~pp loss on corrupted inputs.

\begin{table}[t!]
	\centering
	\scriptsize
	\resizebox{\linewidth}{!}{%
		\begin{tabular}{clr|rrrr}
			\toprule
			Sparsity                      & Method                           & \multicolumn{1}{c}{Active params.} &
			\multicolumn{1}{|c}{$\Acc$}   & \multicolumn{1}{c}{$\Delta\Acc$} &
			\multicolumn{1}{c}{$\DBEacc$} & \multicolumn{1}{c}{$\RobAcc$}                                                                                                \\
			\multicolumn{1}{c}{(\%)}      &                                  & \multicolumn{1}{c}{(millions)}     &
			\multicolumn{1}{|c}{(\%)}     & \multicolumn{1}{c}{(pp)}         &                                    &
			\multicolumn{1}{c}{(\%)}                                                                                                                                     \\
			\midrule
			0                             & Dense                            & 11.17                              & $94.62\pm0.26$ & $0.00$  & $0.000$         & $65.71$ \\
			\addlinespace[1pt]
			\multirow{4}{*}{50}           & LT                               & \multirow{4}{*}{5.59}              & $94.63\pm0.10$ & $+0.01$ & $0.648\pm0.555$ & $65.87$ \\
			                              & Accuracy-only                    &                                    & $94.52\pm0.19$ & $-0.10$ & $0.654\pm0.467$ & $65.37$ \\
			                              & BP-LT                            &                                    & $94.62\pm0.20$ & $0.00$  & $0.446\pm0.304$ & $65.13$ \\
			                              & Random sparse                    &                                    & $94.31\pm0.15$ & $-0.31$ & $0.727\pm0.404$ & $65.34$ \\
			\addlinespace[1pt]
			\multirow{4}{*}{80}           & LT                               & \multirow{4}{*}{2.24}              & $94.46\pm0.17$ & $-0.16$ & $0.501\pm0.248$ & $64.14$ \\
			                              & Accuracy-only                    &                                    & $94.38\pm0.22$ & $-0.24$ & $0.890\pm0.598$ & $64.10$ \\
			                              & BP-LT                            &                                    & $94.27\pm0.51$ & $-0.35$ & $0.437\pm0.168$ & $63.77$ \\
			                              & Random sparse                    &                                    & $93.56\pm0.16$ & $-1.06$ & $0.981\pm0.434$ & $62.95$ \\
			\addlinespace[1pt]
			\multirow{4}{*}{90}           & LT                               & \multirow{4}{*}{1.13}              & $94.13\pm0.06$ & $-0.49$ & $1.364\pm0.455$ & $62.94$ \\
			                              & Accuracy-only                    &                                    & $94.06\pm0.19$ & $-0.56$ & $1.295\pm0.415$ & $63.71$ \\
			                              & BP-LT                            &                                    & $93.85\pm0.20$ & $-0.77$ & $0.995\pm0.156$ & $63.85$ \\
			                              & Random sparse                    &                                    & $92.41\pm0.23$ & $-2.21$ & $1.521\pm0.359$ & $61.52$ \\
			\addlinespace[1pt]
			\multirow{4}{*}{95}           & LT                               & \multirow{4}{*}{0.57}              & $93.38\pm0.10$ & $-1.24$ & $1.059\pm0.095$ & $61.52$ \\
			                              & Accuracy-only                    &                                    & $93.50\pm0.15$ & $-1.12$ & $1.099\pm0.071$ & $61.48$ \\
			                              & BP-LT                            &                                    & $93.28\pm0.20$ & $-1.34$ & $1.014\pm0.138$ & $62.15$ \\
			                              & Random sparse                    &                                    & $90.98\pm0.22$ & $-3.64$ & $2.183\pm0.211$ & $59.05$ \\
			\bottomrule
		\end{tabular}
	}
	\caption{Multi-sparsity results for CIFAR-10 ResNet-18. $\Acc$ is the mean $\pm$ standard deviation across evaluated models, and active parameter counts are in millions. $\Delta\Acc$ is candidate test accuracy minus the matched dense reference's test accuracy (pp), and $\RobAcc$ is corruption accuracy under the setting-specific protocol. This table uses CIFAR-10-C. $\DBEacc$ quantifies the magnitude of the measured behavioral difference from the dense reference, and smaller values indicate closer behavior. These conventions apply to subsequent tables.}
	\label{tab:cifar10-sparsity}
\end{table}

Table~\ref{tab:cifar10-sparsity} shows that LT baselines are strong in clean accuracy at the 50\% and 80\% sparsity levels, yet matching clean accuracy does not establish behavioral compatibility: The gap is visible in OOD response, corruption accuracy, representation distance, and policy behavior. The accuracy constraint also matters. Candidates with global random sparse masks can sometimes have a moderate behavioral-compatibility distance and, in some protocol groups, satisfy the test recovery band, while harder or higher-sparsity groups miss the band. We therefore report $\DBEacc$ together with the explicit accuracy-matching status.

\paragraph{Larger Sparse-Baseline Set at 80\% Sparsity.}

Appendix Table~\ref{tab:large-pool} reports the complementary 50\% large-pool comparison, and Appendix Table~\ref{tab:rigl-80} reports the companion 80\% protocol with RigL and Staged MP. The 50\% comparison shows that behavioral gaps persist with a stronger sparse-baseline set. Because Table~\ref{tab:cifar10-sparsity} already suggests that 80\% sparsity is the stricter sparse-recovery stress point, we evaluate the broader CIFAR-10 ResNet-18 comparison there.

Table~\ref{tab:large-pool-80} is the largest sparse-baseline comparison at 80\% sparsity in this paper. The dense reference's accuracy remains high at $94.53\%$. The LT baseline, LT reinitialization, IMP, SNIP, global random sparse, layerwise random sparse, and one-shot MP protocol--method groups all lie within the $\pm1.5$~pp test recovery band and are thus study-band-matched.

Yet all of these methods show nonzero measured deviations from the dense reference. The LT baseline has $\DBEacc=0.784$, $4.8\%$ prediction disagreement, and $\mathrm{AcceptFlip}_{\tau_D}=8.5\%$. IMP and SNIP recover accuracy but show similar behavioral distances. Candidates with global random sparse masks are weaker in accuracy and farther in behavior. Layerwise random sparse controls and one-shot MP controls have smaller distances but still change measured coordinates relative to the dense reference. BP-LT selection reduces the behavioral distance from $0.784$ for the LT baseline to $0.701$ while preserving the accuracy constraint. BP-LT selection does not reduce the headline policy coordinate in this cell: The mean acceptance-flip rate is $8.54\%$ for BP-LT and $8.47\%$ for LT. This outcome is consistent with BP-LT's role as a multi-coordinate screening rule, not a policy-restoration objective.

The Staged MP protocol--method group represents the recovery-stress stratum of the same setting. Under this staged recipe with a 200-epoch sparse training budget, the Staged MP baseline reaches $91.48\%$, about 3.05~pp below the dense reference. A companion protocol with RigL and Staged MP adopts a lower post-pruning fine-tuning learning rate; results from this separate protocol group support the same replacement audit conclusion that accuracy recovery alone does not certify continuity.

\begin{table}[t!]
	\centering
	\scriptsize
	\resizebox{\linewidth}{!}{%
		\begin{tabular}{l|rrrrrr}
			\toprule
			Method                        & \multicolumn{1}{|c}{$\Acc$}   & \multicolumn{1}{c}{$\Delta\Acc$} &
			\multicolumn{1}{c}{$\DBEacc$} & \multicolumn{1}{c}{OOD}       &
			\multicolumn{1}{c}{Flip}      & \multicolumn{1}{c}{$\RobAcc$}                                                                             \\
			                              & \multicolumn{1}{|c}{(\%)}     & \multicolumn{1}{c}{(pp)}         &         &
			\multicolumn{1}{c}{AUROC}     & \multicolumn{1}{c}{(\%)}      &
			\multicolumn{1}{c}{(\%)}                                                                                                                  \\
			\midrule
			Dense                         & $94.53\pm0.21$                & $0.00$                           & $0.000$ & $0.8609$ & $0.00$  & $72.42$ \\
			\addlinespace[1pt]
			LT                            & $94.39\pm0.05$                & $-0.14$                          & $0.784$ & $0.8427$ & $8.47$  & $71.27$ \\
			LT reinit                     & $94.33\pm0.17$                & $-0.20$                          & $0.784$ & $0.8490$ & $8.87$  & $71.18$ \\
			IMP                           & $94.25\pm0.07$                & $-0.29$                          & $0.758$ & $0.8520$ & $9.02$  & $71.40$ \\
			SNIP                          & $94.31\pm0.09$                & $-0.22$                          & $0.800$ & $0.8554$ & $8.87$  & $70.84$ \\
			Random sparse                 & $93.70\pm0.15$                & $-0.84$                          & $1.047$ & $0.8324$ & $9.98$  & $70.58$ \\
			Layerwise random              & $94.34\pm0.02$                & $-0.19$                          & $0.724$ & $0.8487$ & $8.58$  & $70.79$ \\
			One-shot MP                   & $94.23\pm0.16$                & $-0.31$                          & $0.655$ & $0.8596$ & $6.92$  & $71.86$ \\
			\addlinespace[1pt]
			Staged MP                     & $91.48\pm0.21$                & $-3.05$                          & $1.940$ & $0.8494$ & $15.23$ & $67.28$ \\
			\addlinespace[2pt]
			Accuracy-only                 & $94.30\pm0.09$                & $-0.24$                          & $0.782$ & $0.8447$ & $8.60$  & $71.40$ \\
			BP-LT                         & $94.44\pm0.09$                & $-0.09$                          & $0.701$ & $0.8511$ & $8.54$  & $71.18$ \\
			\addlinespace[1pt]
			Small dense                   & $88.10\pm0.20$                & $-6.44$                          & $2.995$ & $0.7983$ & $21.59$ & $61.01$ \\
			\bottomrule
		\end{tabular}
	}
	\caption{Large-pool CIFAR-10 ResNet-18 at 80\% sparsity. Flip denotes $\mathrm{AcceptFlip}_{\tau_D}$.}
	\label{tab:large-pool-80}
\end{table}

\subsection{Replications and Transfer}

The pattern of accuracy recovery with behavioral gaps recurs across datasets, architectures, and pretrained backbones. Appendix Tables~\ref{tab:replications},~\ref{tab:imagenette50}, and~\ref{tab:transfer} report the full results. On CIFAR-100 ResNet-18 at 50\% sparsity, the LT baseline reaches $74.66\%$, compared with $75.30\%$ for the dense reference, with $\DBEacc=0.768$ and a 1.40~pp $\RobAcc$ loss. CIFAR-10 ResNet-34 at 80\% is within 0.07~pp of the corresponding dense reference but has $\DBEacc=1.029$, and CIFAR-10 WideResNet-28-2 at 80\% stays inside the test recovery band with a 1.17~pp gap, $\DBEacc=1.149$, and a 3.54~pp $\RobAcc$ loss. The pretrained Imagenette protocols at 50\% sparsity extend the audit to modern backbones. The ConvNeXt-Tiny LT baseline reaches $97.94\%$, exceeding the corresponding dense reference's $97.49\%$ clean accuracy, yet still has $\DBEacc=1.205$ and flips $9.69\%$ of dense-threshold decisions. ViT-Tiny recovers to within 0.28~pp with $\DBEacc=0.422$ and $\mathrm{AcceptFlip}_{\tau_D}=8.08\%$. Fine-grained transfer with pretrained ConvNeXt-Tiny repeats the pattern: Flowers-102 at 50\% sparsity recovers to within 0.27~pp with $\DBEacc=0.739$ and $\mathrm{AcceptFlip}_{\tau_D}=7.28\%$, and FGVC-Aircraft at 50\% sparsity recovers to within 0.93~pp with $\DBEacc=0.658$ and $\mathrm{AcceptFlip}_{\tau_D}=7.91\%$. BP-LT lowers the behavioral distance in every one of these settings.

\subsection{Policy-Level Shifts}

\begin{figure}[t!]
	\centering
	\includegraphics[width=0.86\linewidth]{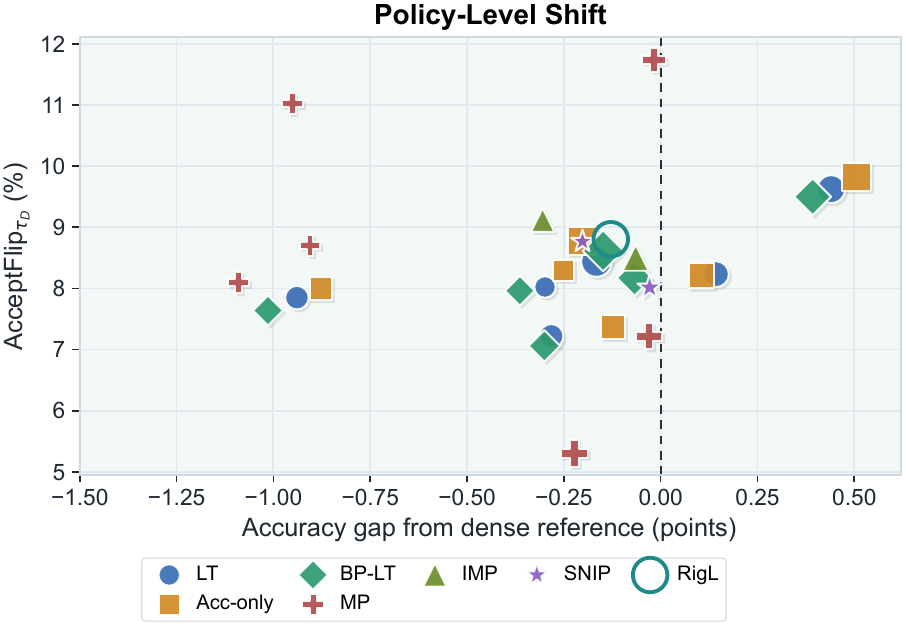}
	\caption{Policy-level shift for small-gap evaluations satisfying $|\Delta\Acc|\leq3$~pp. $\mathrm{AcceptFlip}_{\tau_D}$ is the percentage of test inputs receiving opposite accept--review decisions under the fixed dense threshold. This coordinate pools both directions and is not a top-1 label-flip rate. Markers show grouped method means with offsets for overlaps. Area increases with $\DBEacc$.}
	\label{fig:policy-shift}
\end{figure}

For protocols with fixed-threshold policy diagnostics, Figure~\ref{fig:policy-shift} shows that LT baselines in evaluations with $|\Delta\Acc|\leq3$~pp flip roughly 7\% to 10\% of dense-threshold decisions; grouped method means span 5\% to 12\%. All plotted grouped method means have accuracy gaps within $\pm1.5$~pp, yet candidates with near-zero gaps and those that exceed dense accuracy flip similarly. Accuracy gap therefore does not predict routing churn, consistent with Theorem~\ref{thm:policy-nonidentifiability}, Proposition~\ref{prop:threshold-flip-bound}, and Theorem~\ref{thm:threshold-amplification}. Net accept-to-review crossings can raise review arrivals and latency; review-to-accept crossings bypass incumbent escalation and can alter accepted risk. Offset flows can preserve volume while changing reviewed cases. Acceptance flips thus establish routing non-preservation, not realized cost or harm; quantifying the latter requires directional rates and deployment costs. The Appendix gives full proofs and further interpretation and causal scope.

\section{Conclusion}

Our deployment-perspective evaluation concludes that matching an incumbent dense model's clean accuracy does not ensure that an LT is a policy-compatible replacement at the incumbent's fixed operating point. We show theoretically that the insufficiency of an accuracy-only certificate is structural: Exact top-1 agreement does not identify the dense-threshold policy, and near-threshold confidence movement along regular local paths produces policy flips at a first-order rate governed by the threshold-amplification coefficient. Empirically, across a broad range of settings, sparse candidates that recover the accuracy of the corresponding dense references remain measurably different on active audit coordinates, and the sparse candidates' accept--review changes do not vanish as the accuracy gap approaches zero. Recurring corruption-accuracy losses in study-band-matched settings further show that clean recovery can mask lower deployment-relevant performance, a risk a signed service-level audit can detect. Behavior-aware selection often reduces aggregate compatibility gaps while preserving the accuracy constraint; policy restoration and individual-coordinate requirements remain separate release checks. Strict drop-in claims therefore need paired compatibility checks and signed service-level requirements alongside clean accuracy. For each dense--sparse comparison, we set the threshold from the dense reference's validation confidences to target a 10\% review rate and apply that threshold unchanged to the sparse candidate.

\bibliography{aaai2027}

\clearpage
\appendix

\twocolumn[{%
			\section{Appendix}

			\subsection{Sparse-Replacement Test Overview}

			\begin{minipage}{\textwidth}
				\centering
				\begin{tikzpicture}[
					x=1cm,
					y=1cm,
					>=Latex,
					font=\sffamily\footnotesize,
					panel/.style={fill=lotPanel, draw=lotAxis!18, line width=0.45pt,
							rounded corners=5pt},
					axis/.style={draw=lotAxis, line width=0.8pt, -{Latex[length=2.0mm]}},
					flow/.style={draw=lotAxis!58, line width=0.75pt,
					-{Latex[length=2.0mm]}},
					densept/.style={circle, fill=lotDense, draw=lotPanel, line width=0.55pt,
							minimum size=7.5pt, inner sep=0pt},
					sparsept/.style={circle, fill=lotSparse, draw=lotPanel, line width=0.45pt,
							minimum size=6.8pt, inner sep=0pt},
					bppt/.style={circle, fill=lotBP, draw=lotPanel, line width=0.55pt,
							minimum size=8.0pt, inner sep=0pt},
					policypt/.style={circle, draw=lotPanel, line width=0.42pt,
							minimum size=6.3pt, inner sep=0pt},
					title/.style={font=\sffamily\small\bfseries, text=lotInk},
					sublabel/.style={font=\sffamily\scriptsize\bfseries, text=lotMuted},
					mathlabel/.style={font=\sffamily\scriptsize, text=lotInk,
							inner sep=1.1pt}
					]
					\node[panel, minimum width=5.15cm, minimum height=5.55cm, anchor=south west]
					at (0,0) {};
					\node[panel, minimum width=5.55cm, minimum height=5.55cm, anchor=south west]
					at (5.65,0) {};
					\node[panel, minimum width=5.25cm, minimum height=5.55cm, anchor=south west]
					at (11.55,0) {};
					\draw[flow] (5.18,2.78) -- (5.56,2.78);
					\draw[flow] (11.22,2.78) -- (11.48,2.78);

					\node[title] at (2.58,5.20) {Accuracy Slab};
					\fill[lotBand, rounded corners=2pt, fill opacity=0.86]
					(2.18,0.92) rectangle (3.16,4.28);
					\draw[draw=lotBand!80!lotDense, line width=0.55pt]
					(2.18,0.92) -- (2.18,4.28)
					(3.16,0.92) -- (3.16,4.28);
					\draw[draw=lotDense!75, dashed, line width=0.95pt]
					(2.67,0.76) -- (2.67,4.46);
					\draw[axis] (0.68,0.90) -- (4.72,0.90);
					\draw[axis] (0.68,0.90) -- (0.68,4.28);
					\draw[draw=lotDense!70, line width=0.65pt, {Latex[length=1.6mm]}-{Latex[length=1.6mm]}]
					(2.67,4.55) -- (3.16,4.55);
					\node[mathlabel, text=lotDense, anchor=south] at (2.93,4.58)
					{$\epsilon_{\mathrm{acc}}$};
					\node[mathlabel, anchor=south] at (2.67,4.03)
					{$|\Delta\Acc_{\mathrm{val}}|\leq\epsilon_{\mathrm{acc}}$};
					\node[mathlabel, anchor=north] at (4.54,0.64) {$\Delta\Acc_{\mathrm{val}}$};
					\node[mathlabel, rotate=90] at (0.28,2.58) {$\DBEacc$};
					\foreach \x/\y in {1.30/1.64,1.62/3.24,3.68/3.05,4.02/1.72}
					\node[sparsept, fill=lotSparse!30] at (\x,\y) {};
					\foreach \x/\y in {2.25/3.78,2.93/3.42,2.46/2.88,3.02/2.48,2.22/2.10,2.86/2.02}
					\node[sparsept] at (\x,\y) {};
					\node[densept] (denseA) at (2.67,1.16) {};
					\node[bppt] (bpA) at (2.88,1.72) {};
					\draw[draw=lotSparse!80, dashed, line width=0.9pt]
					(denseA) -- (2.25,3.78);
					\draw[draw=lotBP!90, line width=0.9pt]
					(denseA) -- (bpA);
					\node[mathlabel, text=lotDense, anchor=west] at (2.92,1.10) {$f_D$};
					\node[mathlabel, text=lotBP, anchor=west] at (3.09,1.72) {BP-LT};
					\node[mathlabel, text=lotSparse, anchor=east] at (2.05,2.70) {$\DBEacc$};

					\node[title] at (8.43,5.20) {Behavioral Distance};
					\coordinate (centerB) at (8.43,2.62);
					\fill[lotBP!8] (centerB) circle (0.78);
					\foreach \r in {0.52,1.04,1.56}
					\draw[draw=lotAxis!14, line width=0.62pt] (centerB) circle (\r);
					\draw[draw=lotBP!50, line width=0.82pt] (centerB) circle (0.78);
					\draw[draw=lotAxis!22, line width=0.55pt]
					($(centerB)+(-1.72,0)$) -- ($(centerB)+(1.72,0)$);
					\draw[draw=lotAxis!22, line width=0.55pt]
					($(centerB)+(0,-1.72)$) -- ($(centerB)+(0,1.72)$);

					\foreach \x/\y in {7.28/3.48,7.48/1.68,8.98/4.12,9.92/1.82,10.05/2.52}
					\node[sparsept, fill=lotSparse!32, minimum size=5.5pt] at (\x,\y) {};
					\coordinate (ltB) at (9.70,3.60);
					\coordinate (bpB) at (9.03,2.88);
					\draw[draw=lotSparse!84, dashed, line width=0.92pt]
					(centerB) -- (ltB);
					\draw[draw=lotBP!90, line width=1.02pt]
					(centerB) -- (bpB);
					\node[densept] at (centerB) {};
					\node[sparsept, minimum size=7.5pt] at (ltB) {};
					\node[bppt, minimum size=8.2pt] at (bpB) {};
					\node[mathlabel, text=lotDense, anchor=north east] at (8.33,2.48) {$f_D$};
					\node[mathlabel, text=lotSparse, anchor=south west] at (9.84,3.72) {LT};
					\node[mathlabel, text=lotBP, anchor=west] at (9.28,2.83) {BP-LT};
					\node[mathlabel, text=lotSparse, anchor=south] at (9.05,3.30) {$\DBEacc$};

					\node[title] at (14.18,5.20) {Policy Half-Space};
					\fill[lotPolicy, rounded corners=3pt, fill opacity=0.60]
					(14.36,0.88) rectangle (16.42,4.42);
					\draw[axis] (12.25,3.62) -- (16.20,3.62);
					\draw[axis] (12.25,1.82) -- (16.20,1.82);
					\draw[draw=lotDense!76, dashed, line width=1.0pt]
					(14.36,0.72) -- (14.36,4.48);
					\node[mathlabel, text=lotDense, anchor=south west] at (14.50,4.32) {$\tau_D$};
					\node[sublabel, anchor=west] at (11.78,3.62) {$f_D$};
					\node[sublabel, anchor=west] at (11.78,1.82) {$f$};
					\node[sublabel] at (13.22,4.06) {review};
					\node[sublabel] at (15.46,4.06) {accept};
					\node[mathlabel, anchor=north] at (16.04,0.60) {$c(x)$};

					\draw[draw=lotAxis!26, line width=0.55pt]
					(12.72,3.62) -- (12.88,1.82)
					(13.20,3.62) -- (13.12,1.82)
					(15.36,3.62) -- (15.08,1.82)
					(15.78,3.62) -- (15.56,1.82);
					\draw[draw=lotWarn, line width=0.95pt, -{Latex[length=1.8mm]}]
					(13.82,3.54) .. controls (14.10,2.86) and (14.42,2.36) .. (14.70,1.90);
					\draw[draw=lotWarn, line width=0.95pt, -{Latex[length=1.8mm]}]
					(14.88,3.70) .. controls (14.52,2.88) and (14.18,2.38) .. (13.92,1.90);
					\foreach \x in {12.72,13.20}
					\node[policypt, fill=lotMuted!55] at (\x,3.62) {};
					\foreach \x in {15.36,15.78}
					\node[policypt, fill=lotDense] at (\x,3.62) {};
					\foreach \x in {12.88,13.12}
					\node[policypt, fill=lotMuted!55] at (\x,1.82) {};
					\foreach \x in {15.08,15.56}
					\node[policypt, fill=lotBP] at (\x,1.82) {};
					\node[policypt, fill=lotWarn] at (13.82,3.62) {};
					\node[policypt, fill=lotWarn] at (14.88,3.62) {};
					\node[policypt, fill=lotWarn] at (14.70,1.82) {};
					\node[policypt, fill=lotWarn] at (13.92,1.82) {};
					\node[mathlabel, text=lotWarn, anchor=west] at (14.78,2.64) {flip};
				\end{tikzpicture}
				
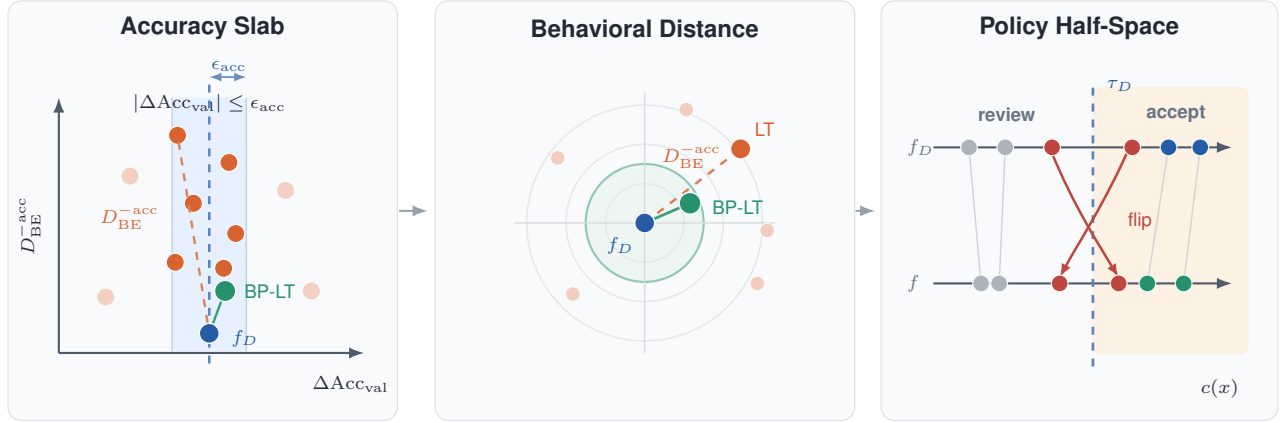
\captionof{figure}{Illustration of the sparse-replacement audit. Sparse candidates first pass through a validation clean-accuracy slab around the dense reference, where $\Delta\Acc_{\mathrm{val}}\coloneqq \SelAcc(f)-\SelAcc(f_D)$. Inside that slab, replacement is screened by clean-accuracy-excluded behavioral-compatibility distance ($\DBEacc$) to the dense reference's behavior vector, and BP-LT selects a closer in-band candidate. The final audit measures changes across the acceptance and review boundary after reusing the dense threshold $\tau_D$.}
				\label{fig:graphical-abstract}
			\end{minipage}
		}]

\subsection{Behavioral Audit Construction}

\subsubsection{List of Notation}

\begin{center}
	\centering
	\scriptsize
	\begin{tabularx}{\linewidth}{>{\raggedright\arraybackslash}X>{\raggedright\arraybackslash}X}
		\toprule
		Symbol                                                                                                                                  & Meaning                                                                                         \\
		\midrule
		$f_D$; $f,f_i$                                                                                                                          & Dense reference; sparse candidates                                                              \\
		$s,\mathcal{F}_s$; $(X,Y),Z$                                                                                                            & Sparsity and candidate pool; clean input--label pair and OOD input                              \\
		$p_f(y\mid x)$; $c_f(x)$; $\hat y_f(x)$                                                                                                 & Predictive distribution; maximum confidence; predicted class                                    \\
		$\tau_D$; $a_f(x),A_f$                                                                                                                  & Dense confidence threshold; acceptance indicator and event                                      \\
		\addlinespace[2pt]
		$\mathbf{b}^{-\mathrm{acc}}$; $\mathbf{b}_{\mathrm{cal}},\mathbf{b}_{\mathrm{OOD}},\mathbf{b}_{\mathrm{pol}},\mathbf{b}_{\mathrm{ref}}$ & Behavior vector and its calibration, OOD, policy, and combined reference and reliability blocks \\
		$\mathrm{AcceptFlip}_{\tau_D}$; $\mathrm{PredDisagree}$; $\RobAcc$                                                                      & Acceptance-policy disagreement; top-1 disagreement; corruption accuracy                         \\
		\addlinespace[2pt]
		$G$, $\mathcal{F}_G$, $\mathcal{M}_G$                                                                                                   & Protocol group; distance-fit rows; active coordinate set                                        \\
		$x_j,\widetilde{x}_j,Z_j$; $d_j,\mu_{j,G},\sigma_{j,G}$                                                                                 & Raw, imputed, and standardized coordinate; direction, mean, and scale                           \\
		$\DBEacc$; $\SelDBE$                                                                                                                    & Test-split and validation-split behavioral-compatibility distances                              \\
		\addlinespace[2pt]
		$D_\tau, \Delta_c$                                                                                                                      & Dense threshold margin and candidate confidence displacement                                    \\
		$V$, $h$, $\kappa_{\tau_D}$                                                                                                             & Local confidence direction, joint density, and threshold-amplification coefficient              \\
		\addlinespace[2pt]
		$\SelAcc,\epsilon_{\mathrm{acc}}$; $\Delta_i^{\sel},\lambda$                                                                            & Validation accuracy and tolerance; normalized gap and penalty weight                            \\
		$f_{\mathrm{BP}},f_{\mathrm{BP}}^{\mathrm{fb}}$; $\Delta\Acc_{\mathrm{val}},\Delta\Acc$                                                 & BP-LT choices; signed validation and test accuracy gaps                                         \\
		$D_{\mathrm{rank}}, D_{\mathrm{Mah}}$                                                                                                   & Sensitivity distances                                                                           \\
		\bottomrule
	\end{tabularx}
	\captionof{table}{Principal notation. Auxiliary symbols and individual behavioral coordinates are defined where introduced.}
	\label{tab:notation}
\end{center}

\subsubsection{Behavioral Coordinates and Distance Details}

\paragraph{Coordinate Roles and Calibration Details.}
The coordinates within a block expose complementary failure modes. ECE measures empirical confidence--accuracy mismatch, while NLL and Brier are proper scoring rules that assess the predictive distribution and penalize confident errors. OOD AUROC measures ranking quality, $\IDFAR$ measures in-distribution false-alarm load at a fixed OOD-recall target, and OOD confidence measures residual confidence on shifted inputs. The policy coordinates translate confidence movement into review volume, accepted error, OOD acceptance, and paired routing changes. Worst-class accuracy detects a class-specific service-level loss hidden by the aggregate. An MIA probes black-box membership exposure, prediction disagreement measures output churn, and corruption accuracy probes performance under the evaluated input shifts. Representation distance is included only as a diagnostic of intermediate-feature compatibility and does not independently establish harm. We compute $\ECE$ with 15 equal-width confidence bins, and $\mathrm{Brier}(f)$ is the multiclass Brier score \citep{naeini2015obtaining,brier1950verification},
\begin{align*}
	\mathrm{Brier}(f)
	 & \coloneqq
	\E[\sum_{y=1}^{K}(p_f(y\mid X)-\mathbf{1}\{Y=y\})^2].
\end{align*}
We use $\NLL(f)\coloneqq\E[-\log p_f(Y\mid X)]$ on the same clean evaluation law. We also use $\mathrm{WorstClassAcc}(f)\coloneqq \min_{y:\Pr[Y=y]>0}\Pr[\hat y_f(X)=y\mid Y=y]$.

The following paragraphs provide detailed definitions of the OOD, privacy, representation, prediction-disagreement, and corruption coordinates, as well as the distance-fit, normalization, active-coordinate, missing-value, and scale-fallback rules used below. In particular, $\mathrm{PredDisagree}$ measures top-1 disagreement with the dense reference, and $\RobAcc$ measures corruption accuracy on a corruption distribution.

\paragraph{OOD Coordinates.}
Let $p_f(y\mid x)$ be the predictive distribution, $c_f(x)\coloneqq\max_y p_f(y\mid x)$, and $\hat y_f(x)\coloneqq\arg\max_y p_f(y\mid x)$ under a fixed measurable tie-breaking rule. For the OOD coordinates, let $s_f(x)\coloneqq-c_f(x)$ be the OOD score, so larger values indicate more OOD-like examples. Clean accuracy is $\Acc(f)\coloneqq\Pr[\hat y_f(X)=Y]$. Then $\AUROC_{\mathrm{OOD}}(f)$ is computed for the binary label $\mathbf{1}\{\mathrm{OOD}\}$ with score $s_f$. The orientation-specific in-distribution false-alarm coordinate is $\IDFAR(f)\coloneqq \Pr[s_f(X)\ge q_{0.05}(s_f(Z))]$, where $q_{0.05}$ is the 5\% quantile of the OOD-score distribution. Thus, this coordinate measures in-distribution false alarms at a nominal OOD recall of 95\%, induced by this quantile rule. With finite samples and score ties, $\IDFAR$ is the corresponding empirical quantile diagnostic under the OOD-positive scoring convention. The $\IDFAR$ coordinate is distinct from the conventional OOD-detection false-positive rate at a 95\% true-positive rate, which treats in-distribution examples as positive and reports OOD acceptance when the in-distribution true-positive rate is 95\%. Under the same maximum-confidence score, the latter is $\Pr[c_f(Z)\geq q_{0.05}(c_f(X))]$ up to the empirical tie convention. Finally, $\mathrm{Conf}_{\mathrm{OOD}}(f)\coloneqq\E[c_f(Z)]$.

\paragraph{Policy-Coordinate Direction and Missing Values.}
The two directions of this disagreement are
\begin{align*}
	r_{\mathrm{A}\to\mathrm{R}}(f,f_D)
	 & \coloneqq\Pr[a_D(X)=1, a_f(X)=0], \\
	r_{\mathrm{R}\to\mathrm{A}}(f,f_D)
	 & \coloneqq\Pr[a_D(X)=0, a_f(X)=1].
\end{align*}
Suppressing the shared arguments, the two directional rates satisfy
\begin{align*}
	\mathrm{AcceptFlip}_{\tau_D}
	 & =r_{\mathrm{A}\to\mathrm{R}}+r_{\mathrm{R}\to\mathrm{A}}, \\
	\Delta\mathrm{ReviewRate}_{\tau_D}
	 & =r_{\mathrm{A}\to\mathrm{R}}-r_{\mathrm{R}\to\mathrm{A}}.
\end{align*}
Here $\Delta\mathrm{ReviewRate}_{\tau_D}$ denotes the candidate rate minus the dense-reference rate. Thus, opposite-direction changes can cancel in the aggregate review rate even when many individual inputs are routed differently.

The $\mathrm{AutoErr}_{\tau_D}$ coordinate is defined when $\Pr[A_f]>0$. If $\Pr[A_f]=0$, the coordinate is undefined and reported as missing. Thus, $\mathrm{AutoErr}_{\tau_D}$ is the conditional error rate among automatically accepted in-distribution examples, and $\mathrm{AcceptFlip}_{\tau_D}$ is the fraction of in-distribution examples whose acceptance or review decision changes relative to the dense reference.

\paragraph{Policy-Coordinate Interpretation.}
Here, review denotes operational abstention or escalation to a human or a downstream verification system; the experiments audit routing decisions and do not assume a particular reviewer or measure the reviewer's outcome. We also report prediction disagreement with the dense reference. For a service processing $N$ inputs, an acceptance-flip rate $\phi$ reroutes approximately $\phi N$ inputs even when opposite flip directions leave the net review rate nearly unchanged.

$\mathrm{AcceptFlip}_{\tau_D}$ measures a different object from a continuous confidence difference. A large confidence change that remains on the same side of $\tau_D$ leaves the service action unchanged, whereas an arbitrarily small change across $\tau_D$ changes whether that input is automatically accepted or reviewed. The flip rate therefore directly tests compatibility with the fixed deployed gate and complements, rather than replaces, calibration and confidence diagnostics. The two flip directions also have different operational implications: An accept-to-review change can increase review workload and latency, while a review-to-accept change bypasses an escalation required by the dense policy and can change risk exposure. Because $\mathrm{AcceptFlip}_{\tau_D}$ pools these directions, the acceptance-flip rate is a bidirectional policy-equivalence diagnostic rather than a signed utility measure. Together, the four policy coordinates---$\mathrm{ReviewRate}_{\tau_D}$, $\mathrm{AutoErr}_{\tau_D}$, $\mathrm{OODAccept}_{\tau_D}$, and $\mathrm{AcceptFlip}_{\tau_D}$---test whether sparse replacement preserves the operating point of a selective-classification or OOD-filtering policy configured for the dense reference and characterize the consequences of failing to preserve that operating point.

\paragraph{Privacy Coordinate.}
The $\MIA(f)$ coordinate is a compact black-box MIA score reported on the AUC-excess scale. This choice follows the overfitting-based privacy-risk view developed by \citet{yeom2018privacy}, the black-box attack setting introduced by \citet{shokri2017membership}, and later attack-score refinements \citep{carlini2022membership}:
\begin{align*}
	A_{\mathrm{score}}(f)
	 & \coloneqq
	\max_{s\in\mathcal{S}_f} a_{\mathrm{AUC}}(s),                        \\
	a_{\mathrm{AUC}}(s)
	 & \coloneqq
	\max\{\AUC_{\mathrm{mem}}(s),1-\AUC_{\mathrm{mem}}(s)\}-\frac{1}{2}, \\
	\mathcal{S}_f
	 & \coloneqq\{(x,y)\mapsto c_f(x),                                   \\
	 & \hphantom{\coloneqq\{}(x,y)\mapsto-H(p_f(\cdot\mid x)),           \\
	 & \hphantom{\coloneqq\{}(x,y)\mapsto-\ell_Y(f;x,y)\}.
\end{align*}
The reported privacy coordinate is
\begin{align*}
	\MIA(f)
	 & \coloneqq
	\begin{cases}
		\max\{A_{\mathrm{score}}(f),A_{\mathrm{lr}}(f)\},
		                       & A_{\mathrm{lr}}(f)\in\mathbb{R}, \\
		A_{\mathrm{score}}(f), & \text{otherwise},
	\end{cases}
\end{align*}
where $H$ is predictive entropy, $\ell_Y(f;X,Y)$ is target-label loss, and $A_{\mathrm{lr}}$ is the held-out logistic-regression black-box attack measured on the same AUC-excess scale. Here $\AUC_{\mathrm{mem}}$ is computed using the binary membership label $\mathbf{1}\{\mathrm{member}\}$, with training examples treated as members and test examples as nonmembers. The $\MIA$ coordinate is an AUC excess in $[0,1/2]$, not a doubled advantage.

\paragraph{Representation and Corruption.}
We use linear centered kernel alignment (CKA) on penultimate representations \citep{kornblith2019similarity} and set $\RepDist(f,f_D)\coloneqq1-\mathrm{CKA}(f,f_D)$ as the representation-dissimilarity coordinate. We also use $\mathrm{PredDisagree}(f,f_D)\coloneqq \Pr[\hat y_f(X)\neq\hat y_{f_D}(X)]$. $\RobAcc$ is corruption accuracy on a corruption distribution.

\paragraph{Behavioral-Compatibility Distance Details.}
Let $G$ be the relevant protocol group and let $\mathcal{F}_G$ be the set of distance-fit rows. These rows are designated by the analysis protocol for estimating coordinate normalization statistics within that group and do not form an additional model class. The designated rows are fixed before scoring candidates and are never borrowed from a different protocol group. In the reported artifact, $\mathcal{F}_G$ contains the available dense, small dense, one-shot MP, Staged MP, global random sparse, layerwise random sparse, LT reinitialization, IMP, SNIP, Gradient Signal Preservation (GraSP), Iterative Synaptic Flow Pruning (SynFlow), and RigL rows. LT-baseline and validation-selection rows are not used to fit these normalization statistics. A protocol group is the set of rows sharing the dense-reference identity and seed, dataset, architecture, sparsity, training recipe, candidate-pool, and behavioral-evaluation keys that affect the comparison.

For coordinate $j$, let $x_j(f)$ be the raw candidate value, $d_j\in\{-1,1\}$ be the utility direction of coordinate $j$, and let $\mu_{j,G}$ and $\sigma_{j,G}$ be the population mean and standard deviation of the finite values over $\mathcal{F}_G$. In the implemented distance, $\sigma_{j,G}\coloneqq1$ when the computed scale is nonfinite or smaller than $10^{-8}$, including the case of a single finite fit value. Define the active coordinate set
\begin{align*}
	\mathcal{M}_G\coloneqq
	\{j: & x_j(f_D) \text{ is finite and}                         \\
	     & x_j(g) \text{ is finite for some }g\in\mathcal{F}_G\}.
\end{align*}
For $j\in\mathcal{M}_G$, set
\begin{align*}
	\widetilde{x}_j(f)\coloneqq
	\begin{cases}
		x_j(f),    & x_j(f) \text{ is finite}, \\
		\mu_{j,G}, & \text{otherwise},
	\end{cases}
\end{align*}
and
\begin{align*}
	Z_j(f)
	 & \coloneqq
	d_j\frac{\widetilde{x}_j(f)-\mu_{j,G}}{\sigma_{j,G}} .
\end{align*}
Together, these active standardized coordinates form the vectors compared by $\DBEacc$ in the final replacement audit and by the validation analogue $\SelDBE$ in BP-LT selection.

The default equal-coordinate root-mean-square is designed for comparative screening within a protocol. Because the normalization statistics and the active coordinate set are protocol specific, the summary is interpreted primarily within a protocol, and cross-protocol magnitude comparisons are qualitative. The operationally interpretable evidence in our results comes from raw paired policy changes and signed service-level coordinates; the distance summarizes multi-coordinate deviation for within-protocol screening and selection.

\subsection{Theoretical Results}

\subsubsection{Proof of Theorem~\ref{thm:policy-nonidentifiability}}

\paragraph{Proof.}
Because $P_X$ is nonatomic, for any $\alpha\in[0,1]$ there is a measurable set $T_\alpha$ with $P_X(T_\alpha)=\alpha$. Define the desired candidate acceptance set by the symmetric difference
\begin{align*}
	S_\alpha\coloneqq A_D\mathbin{\triangle}T_\alpha.
\end{align*}
Choose constants $c_-\in(1/K,\tau_D)$ and $c_+\in(\tau_D,1)$, and set
\begin{align*}
	c_\alpha(x)
	 & \coloneqq
	\begin{cases}
		c_+, & x\in S_\alpha,    \\
		c_-, & x\notin S_\alpha.
	\end{cases}
\end{align*}
For $k_D(x)\coloneqq\hat y_{f_D}(x)$, define a candidate predictive distribution by
\begin{align*}
	p_{f_\alpha}(y\mid x)
	 & \coloneqq
	\begin{cases}
		c_\alpha(x),                & y=k_D(x),     \\
		\dfrac{1-c_\alpha(x)}{K-1}, & y\neq k_D(x).
	\end{cases}
\end{align*}
These measurable, nonnegative probabilities sum to one. Since $c_\alpha(x)>1/K$, the designated class is the unique maximizer, so $\hat y_{f_\alpha}(x)=\hat y_{f_D}(x)$ for every $x$. The pointwise prediction identity implies equal top-1 accuracy and zero prediction disagreement. The maximum confidence of $f_\alpha$ is $c_\alpha(x)$, so the candidate's acceptance set is exactly $S_\alpha$. Therefore
\begin{align*}
	\mathrm{AcceptFlip}_{\tau_D}(f_\alpha,f_D)
	 & =P_X(S_\alpha\mathbin{\triangle}A_D) \\
	 & =P_X(T_\alpha)=\alpha.
\end{align*}
The construction proves the theorem.
\hfill$\square$

\subsubsection{Proof of Proposition~\ref{prop:threshold-flip-bound}}

\paragraph{Proof.}
The dense policy accepts exactly when $D_\tau\geq0$, whereas the candidate accepts exactly when $D_\tau+\Delta_c\geq0$. The two disjoint ways in which these indicators differ give the equality in Eq.~\ref{eq:threshold-flip-sandwich}. On either event, $|\Delta_c|\geq|D_\tau|$; equality can occur in the second event when the candidate lies exactly at the threshold. Hence, for every $\varepsilon>0$, a flip implies either $|D_\tau|\leq\varepsilon$ or $|\Delta_c|>\varepsilon$. Taking probabilities proves the stated bound. Markov's inequality gives
\begin{align*}
	\Pr[|\Delta_c|>\varepsilon]
	\leq\frac{\E|\Delta_c|^p}{\varepsilon^p}.
\end{align*}
The preceding bounds prove the proposition.
\hfill$\square$

\subsubsection{Proof of Theorem~\ref{thm:threshold-amplification}}

\paragraph{Proof.}
Clipping to $[1/K,1]$ preserves the side of the interior threshold $\tau_D$ on which the perturbed score lies. Write $d=D_\tau$ and partition the joint-density integral by the sign of $v$. Because the joint law has a density, all boundary events corresponding to the interval endpoints below have probability zero. When $v>0$, a flip occurs for $-tv\leq d<0$; when $v<0$, a flip occurs for $0\leq d<-tv$. The joint-density assumption therefore gives
\begin{align*}
	p_t
	 & \coloneqq\mathrm{AcceptFlip}_{\tau_D}(f_t,f_D)        \\
	 & =\int_{v>0}\int_{-tv}^{0}h(d,v)\mathrm{d}d\mathrm{d}v
	+\int_{v<0}\int_{0}^{-tv}h(d,v)\mathrm{d}d\mathrm{d}v.
\end{align*}
After substituting $d=tu$ and dividing by $t$,
\begin{align*}
	\frac{p_t}{t}
	 & =\int_{v>0}\int_{-v}^{0}h(tu,v)\mathrm{d}u\mathrm{d}v  \\
	 & +\int_{v<0}\int_{0}^{-v}h(tu,v)\mathrm{d}u\mathrm{d}v.
\end{align*}
Choose $M<\infty$ with $|V|\leq M$ almost surely. The density vanishes almost everywhere outside $|v|\leq M$, so the integrals may be restricted to that set. For all sufficiently small $t$ satisfying $tM\leq d_0$, continuity at $d=0$ makes each inner integral converge to the corresponding interval length times $h(0,v)$ for almost every $v$. Moreover, each inner integral is bounded by $|v|\bar h(v)$, which is integrable by assumption. Dominated convergence in $v$ therefore yields
\begin{align*}
	\lim_{t\to0^+}\frac{p_t}{t}
	 & =\int_{v>0}v h(0,v)\mathrm{d}v
	+\int_{v<0}(-v)h(0,v)\mathrm{d}v           \\
	 & =\int_{\mathbb{R}}|v|h(0,v)\mathrm{d}v.
\end{align*}
The dominated convergence argument completes the proof.
\hfill$\square$

\subsubsection{Interpretation of the Theoretical Results}

The first theorem establishes a structural limitation of accuracy-only certification; the proposition and local theorem then turn that limitation into a candidate-level, testable account of when threshold flips occur. Clean accuracy records only whether the class with the largest predicted probability is correct, whereas the deployed gate tests whether that largest probability is above $\tau_D$. Theorem~\ref{thm:policy-nonidentifiability} therefore says that, for every desired flip rate, one can construct a candidate that keeps every top-1 prediction unchanged yet reverses the accept--review status of that fraction of inputs. The ability to induce an arbitrary flip rate establishes an information limit of the accuracy certificate; empirical candidates are characterized by the following finite-perturbation analysis. Proposition~\ref{prop:threshold-flip-bound} then localizes the risk for a particular candidate: For any chosen tolerance $\varepsilon$, every flip must come either from an input whose dense confidence was already within $\varepsilon$ of the threshold or from a confidence change larger than $\varepsilon$. Thus, even uniformly small confidence changes can matter when many inputs are crowded near the deployed threshold.

Theorem~\ref{thm:threshold-amplification} makes this boundary effect quantitative along a regular local path. For a perturbation of scale $t$, the flip rate is $t\kappa_{\tau_D}+o(t)$; the threshold-amplification coefficient is large when the dense-confidence distribution has high density at the threshold and the perturbation produces large confidence changes for examples near that boundary. The coefficient is a local sensitivity coefficient, while the finite flip rates from 7\% to 10\% in our experiments are direct empirical measurements. Operationally, an accuracy-recovery check captures neither this boundary sensitivity nor the tail of larger confidence changes. A policy that reuses a fixed threshold inherited from the dense reference must therefore be re-audited after sparse replacement, motivating the reported $\mathrm{AcceptFlip}_{\tau_D}$ coordinate and behavior-aware selection within the accuracy-matched candidate set.

\subsection{Behavior-Aware Sparse-Candidate Screening}

To explore whether the conventional LT recovery target can be extended from clean-accuracy matching to closer agreement with the dense incumbent's measured behavior, we design and evaluate the BP-LT procedure as a validation-time search over a fixed pool of trained sparse candidates. This experiment tests whether behavior-aware selection can identify a more compatible replacement after accuracy recovery. The main text specifies the validation-only selection setup, primary BP-LT in-band selection rule, and parameter values. The subsections below state the scope of this selection unit and specify the pool contents, candidate eligibility, and restricted-pool variants.

\subsubsection{Selection Setup}

\paragraph{Validation-Only Selection Procedure.}
For each matched selection cell, we first compute the dense and candidate clean accuracies on held-out validation data and compute $\SelDBE$ from the active behavioral coordinates and held-out diagnostic sources available to that protocol. The normalized validation accuracy gap and validation band are defined in the main text. We then determine which candidates fall inside the band and apply the selection rules below; each rule selects one candidate. Finally, we recompute the reported metrics on the test split. No test metric is used to choose a candidate. The validation band used for selection and the equally wide test recovery band used for final evidence are distinct: Passing the former does not guarantee passing the latter, so test recovery status is reported separately.

\subsubsection{Selection Rules}

\paragraph{Accuracy-Only Selection.}
The accuracy-only baseline chooses the eligible candidate whose finite validation accuracy is closest to that of the dense reference. The validation band determines whether this selection is in-band; if no candidate is in-band, the closest-accuracy selection is flagged as a fallback and the test accuracy gap of the selected candidate is reported.

\paragraph{Parameter Settings and Units.}
Because selection accuracy is represented as a fraction, $0.015$ corresponds to $1.5$~pp. Throughout, we use \% for percentage-valued levels or rates and pp for differences between percentage-valued quantities. These common values define the reported operating rule rather than universal deployment constants; applications can set the band and trade-off to match application-specific replacement contracts.

\subsubsection{Scope and Selection Unit}

BP-LT serves as shorthand for behavior-aware validation-time screening of already trained sparse candidates. The default eligible pool can include non-LT sparse methods, and selection is performed with respect to the aggregate compatibility objective; ticket construction, retraining, and coordinate-wise release checks remain separate stages. Selection is performed independently within each matched dense-reference seed, sparsity, and protocol cell. Thus, candidates associated with different dense references, seeds, or training and behavioral-evaluation protocols are never pooled into the same selection problem.

\subsubsection{Candidate Pool and Eligibility}

For sparsity $s$, let $\mathcal{F}_s\coloneqq\{f_i\}_{i=1}^n$ denote the fixed eligible pool for one such matched selection cell. BP-LT selects one existing model from this pool. By default, BP-LT uses the protocol-eligible all-sparse pool. Here, all-sparse means all alternatives designated as selection eligible by that protocol, not every sparse baseline reported in the paper. The standard pools contain LT, LT reinitialization, global random, and layerwise random candidates; larger protocols add eligible IMP, SNIP, or RigL candidates as specified in the experimental protocol. One-shot MP and Staged MP are reported as separate post-training controls but are not selection eligible, and neither the dense reference nor the small dense control belongs to $\mathcal{F}_s$. The pool and eligibility flags, including the restricted-pool variants below, are fixed before candidate scoring, and the selected source is retained in the accompanying metadata.

We also evaluate two restricted variants. Ticket-family BP-LT restricts the available candidates to LT, LT reinitialization, IMP, and SNIP when those sources are present, whereas LT-only BP-LT selects only from measured LT-baseline candidates. These variants distinguish gains from broad sparse screening from gains available within progressively narrower ticket pools.

\subsubsection{Fallback and Edge-Case Rules}

If the validation accuracy band is empty, BP-LT instead selects
\begin{align*}
	f_{\mathrm{BP}}^{\mathrm{fb}}
	 & \in
	\operatorname*{argmin}_{f_i\in\mathcal{F}_s}
		[
			\Delta_i^{\sel}
			+
			\lambda
			(\SelDBE(f_i,f_D))^2
		],
\end{align*}
with deterministic tie-breaking. In the in-band case, ties prefer smaller behavioral distance and then smaller accuracy gap. In the out-of-band case, ties prefer smaller accuracy gap and then smaller behavioral distance. Out-of-band selections are flagged and remain tied to the same replacement criterion, with test accuracy gaps reported explicitly. For BP-LT, candidates with a nonfinite $\SelAcc$ or $\SelDBE$ are ineligible for these argmin operations. If the dense validation accuracy is nonfinite or no eligible candidate remains, the selection is undefined.

\subsubsection{BP-LT Selection Implementation Excerpt}

Listing~\ref{lst:bp-lt-selection} gives the core validation-time selection rule used for BP-LT. The implementation also handles row metadata and restricted candidate pools. The scoring logic selects inside the validation accuracy band when possible and, when the band is empty, minimizes the normalized accuracy gap plus the weighted squared behavioral distance shown above. In the excerpt, test accuracy is never substituted for missing or nonfinite validation accuracy, keeping candidate selection independent of the test split.

\begin{lstlisting}[
  style=pythonlisting,
  basicstyle=\ttfamily\fontsize{7pt}{7.5pt}\selectfont,
  caption={Python implementation example of BP-LT validation-time selection. Each row
  stores validation accuracy \texttt{sel\_acc}, validation behavioral-compatibility
  distance \texttt{sel\_d\_be\_no\_acc}, and a
  \texttt{candidate\_for\_selection} flag.},
  label={lst:bp-lt-selection}
]
import math


def bp_lt_select(rows, dense, acc_band, lam=0.2):
    def acc(row):
        return float(row.get("sel_acc", math.nan))

    def distance(row):
        value = float(row.get("sel_d_be_no_acc", math.inf))
        return value if math.isfinite(value) else math.inf

    dense_acc = acc(dense)
    if not math.isfinite(dense_acc):
        return None

    candidates = [
        r for r in rows
        if bool(r.get("candidate_for_selection", False))
        and math.isfinite(acc(r))
        and math.isfinite(distance(r))
    ]
    if not candidates:
        return None

    def gap(row):
        value = abs(acc(row) - dense_acc)
        return value if math.isfinite(value) else math.inf

    in_band = [r for r in candidates if gap(r) <= acc_band]
    pool = in_band or candidates
    band = max(float(acc_band), 1e-12)

    def score(row):
        g = gap(row)
        d = distance(row)
        if in_band:
            return (d + lam * (g / band) ** 2, d, g)
        return (g / band + lam * d ** 2, g, d)

    best = min(pool, key=score)
    selected = dict(best)
    selected["method"] = "bp_lt"
    selected["selected_from"] = best.get("method")
    rule_in_band = (
        "validation_acc_band_plus_"
        "dense_behavior_distance"
    )
    rule_fallback = (
        "fallback_acc_behavior_"
        "no_in_band"
    )
    selected["selection_rule"] = (
        rule_in_band if in_band else rule_fallback
    )
    selected["candidate_for_selection"] = False
    return selected
\end{lstlisting}

\subsection{Experimental Protocol and Reproducibility}
\subsubsection{Protocol Definitions, Inclusion, and Recovery Criteria}

An experiment group is the aggregation unit. A protocol group is the behavioral-distance computation unit, defined by a unique combination of dense reference, dense-reference seed, dataset, architecture, sparsity, training recipe, candidate pool, and behavioral-evaluation choices. A selection group is the corresponding validation-time selection problem for a fixed dense reference and selection-eligible candidate pool, so selection-group counts need not equal experiment-group counts. A protocol--method group is a protocol group further split by the reported method.

We treat candidate-pool breadth, corruption source, pruning-baseline budget, learning-rate overrides, auxiliary split policy, selection-eligibility pool, and training and evaluation limits as protocol keys. Using these protocol keys keeps specialized protocols with stronger baselines in the corresponding protocol groups. Hardware, group definitions and required-run inclusion criteria, exact dataset--architecture--sparsity cells, candidate eligibility and counts, training recipes, auxiliary-data coverage, and statistical summaries are fixed by the reported protocol before model aggregation.

We verify compliance with prespecified experiment-group inclusion criteria before aggregation. Within each experiment group, every required seed--sparsity cell must contain the dense reference, one-shot MP, an LT-baseline candidate, and variants produced by the validation selection rules, together with any protocol-specific strong baselines such as SNIP, IMP, RigL, or Staged MP. These criteria concern the availability of required model rows rather than metric values or recovery outcomes; all 56 experiment groups satisfy the inclusion criteria.

We stratify each protocol--method group by recovery quality. A group provides study-band-matched evidence under our strict study-defined criterion when the group's dense reference passes the dataset-specific accuracy floor and every evaluated accuracy gap in the group is finite and lies within the $\pm1.5$~pp test recovery band. Here, the strict label means that every evaluated gap in the protocol--method group satisfies the rule; raw gaps remain reported so that narrower application-specific margins can be applied. This test recovery band differs in both role and data split from the validation accuracy band used by the selection rules. Both bands have a 1.5~pp tolerance, but we use the former to classify test evidence and the latter to screen candidates on validation data. This $\pm1.5$~pp band is a prespecified stratification convention, not a universal equivalence margin or an application release tolerance. The dense-reference accuracy floors are 93\% for CIFAR-10, 72\% for CIFAR-100, 85\% for Imagenette, 90\% for Flowers-102, and 75\% for FGVC-Aircraft. We place a group in the near-recovery stratum if the group's dense reference passes the accuracy floor, the group fails this strict criterion, and the group's absolute mean accuracy gap is at most 3~pp. Other evaluated sparse protocol--method groups that fail the dense-reference accuracy floor or the remaining recovery criteria form recovery-stress regimes. These regimes characterize how behavioral gaps evolve as sparse recovery degrades. This stratification separates the accuracy-matched replacement audit from the transition regimes where sparse-recovery quality changes.

The term accuracy-matched below denotes a regime or setting that satisfies this study-defined rule; the term is not an application-independent equivalence claim. Near-recovery and recovery-stress regimes are identified explicitly. We base policy-level statements on evaluations with the corresponding policy coordinates. We compute aggregate behavioral distances from coordinates with finite dense-reference values and at least one finite distance-fit value in the protocol group. Active coordinates missing for a candidate are imputed with the protocol distance-fit mean. Diagnostics unmeasured at the protocol level remain inactive for that protocol.

\subsubsection{Datasets, Architectures, and Evaluation Sources}

We conduct the primary evaluation on CIFAR-10 with a CIFAR-style ResNet-18 \citep{krizhevsky2009learning,he2016deep} at 50\%, 80\%, 90\%, and 95\% sparsity. We also evaluate CIFAR-100 with ResNet-18 at 50\% and 80\% sparsity, CIFAR-10 with ResNet-34 at 80\% sparsity, and CIFAR-10 with WideResNet-28-2 \citep{zagoruyko2016wide} at 80\% sparsity for architecture replication; replication counts for these settings are reported in the Seed and Candidate-Pool Details subsection below.

The standard CIFAR protocols use SVHN \citep{netzer2011reading} as the OOD dataset. The same experimental setup also supports the 10-class self-taught learning dataset (STL-10) \citep{coates2011analysis}, CIFAR-100, and directory-structured OOD datasets such as Tiny-ImageNet. The large-pool CIFAR-10 protocol merges a broader OOD pool from SVHN, CIFAR-100, and STL-10 with all 15 corruption families in CIFAR-10-C at severities 1, 3, and 5 under a 10,000-example evaluation cap. CIFAR-10-C corruptions are used for CIFAR-10 corruption-accuracy evaluation when available.

Imagenette is an ImageNet-derived, higher-resolution image-classification benchmark \citep{deng2009imagenet}. We use Imagenette for ConvNeXt-Tiny \citep{liu2022convnet} and ViT-Tiny \citep{dosovitskiy2021image} at 50\% sparsity with pretrained PyTorch Image Models (timm) backbones \citep{rw2019timm} and no small dense baseline. The OOD and corruption selection splits are disjoint from the corresponding evaluation splits. These pretrained and transfer settings follow evidence that LT structure can persist after vision pretraining and that sparse ImageNet models can transfer to downstream tasks \citep{chen2021lottery,iofinova2022sparse}.

The fine-grained transfer suite contains three evaluations of pretrained ConvNeXt-Tiny models on 224-pixel inputs. Flowers-102 \citep{nilsback2008automated} at 80\% sparsity is a recovery-stress setting. Flowers-102 at 50\% sparsity is a transfer protocol without a small dense baseline. FGVC-Aircraft \citep{maji2013fine} at 50\% sparsity is a transfer protocol. Auxiliary split status is part of the protocol key for transfer experiments. The Flowers-102 50\% protocol is marked as disjoint, whereas the FGVC-Aircraft protocol has no recorded disjoint-split designation and is analyzed separately.

\subsubsection{Base Training Recipe}
Comparisons use protocol-matched training budgets in both from-scratch and pretrained regimes. The dense references and candidates from the LT baseline, LT reinitialization, global random sparse, and layerwise random sparse methods are trained from scratch for 200 epochs with stochastic gradient descent, a base learning rate of $1\times10^{-1}$, momentum 0.9, Nesterov updates, weight decay $5\times10^{-4}$, random cropping and flipping, a batch size of 512, bfloat16 autocast, a 5-epoch warmup, and cosine learning-rate decay. One-shot MP candidates start from the trained dense reference, apply the one-shot global MP mask, and fine-tune for 30 epochs on CIFAR-10 and 40 epochs on CIFAR-100.

We report active parameter counts for sparse networks. CIFAR-10 ResNet-18 has 11.17 million dense parameters and 5.59, 2.24, 1.13, and 0.57 million active parameters at 50\%, 80\%, 90\%, and 95\% sparsity. CIFAR-10 WideResNet-28-2 has 1.47 million dense parameters and 0.30 million active parameters at 80\% sparsity.

Following a common convention in unstructured weight pruning \citep{gale2019state,tanaka2020pruning}, pruning masks are applied only to trainable convolutional and linear weight tensors. Biases, normalization parameters, and non-weight state are left dense; this convention is important for transformer experiments, where positional embeddings and class tokens are kept outside the ordinary prunable weight matrices.

Global MP, IMP, SNIP, GraSP, and SynFlow masks are constructed by selecting the exact global top-$k$ flattened indices from the pruning scores, rather than by thresholding at the cutoff. This exact top-$k$ selection prevents score ties, which can occur during pruning at initialization, from silently changing the requested number of active maskable weights.

When included, the RigL-style baseline is trained with the same sparse training budget as other sparse candidates trained from initialization. Protocol groups distinguish both the presence of this baseline and the number of topology updates actually performed.

\subsubsection{Modern and Transfer Recipes}
For the Imagenette protocols with 224-pixel inputs, the ConvNeXt-Tiny and ViT-Tiny backbones are initialized with pretrained weights provided by timm, while both classifier heads are initialized randomly. Both models are then trained for 100 epochs using AdamW with a learning rate of $2\times10^{-4}$, weight decay $5\times10^{-2}$, label smoothing $0.1$, exclusion of normalization parameters and biases from weight decay, gradient clipping at 1.0, and cosine learning-rate decay. The LT rewinding point consists of the pretrained backbone and the classifier head's initial random parameter values. The Imagenette OOD pool is a disjoint union of CIFAR-100 and Tiny-ImageNet splits, resized to 224 pixels and normalized with ImageNet statistics; corruption accuracy is evaluated under deterministic Gaussian noise applied to held-out Imagenette test and validation images.

The Flowers-102 and FGVC-Aircraft transfer protocols follow the same pretrained ConvNeXt-Tiny recipe as the Imagenette protocol, with 100 epochs for dense references and sparse candidates and a 20-epoch MP fine-tuning run.

\subsubsection{Candidate Pools and Sparse Baselines}

\paragraph{Baseline definitions and selection eligibility.}
Throughout the experiments, an LT baseline is a sparse candidate obtained by applying the associated LT mask and training the resulting subnetwork from the protocol-specific rewinding point. The rewinding point is the original random initialization in from-scratch protocols or the pretrained backbone with the classifier head restored to the initially sampled random parameter values in pretrained protocols. Tables label these rows LT. LT reinitialization trains the same LT mask from a separately sampled random initialization; tables abbreviate this method as LT reinit.

One-shot MP denotes post-training global MP followed by fine-tuning; tables label this baseline One-shot MP. Staged MP denotes the separate multi-stage post-training MP baseline. The labels Random sparse and Layerwise random denote, respectively, global random sparse masks and random masks that preserve the corresponding LT mask's per-layer sparsity.

Small dense denotes a fully dense, fixed-width convolutional network with base width 48, trained using the protocol's dataset-specific recipe. The Small dense model is a lower-capacity baseline rather than a sparsified version of the dense reference; tables label this row Small dense.

Candidate eligibility is distinct from baseline reporting. The default all-sparse pool comprises the sparse alternatives marked selection eligible for each protocol, and larger protocols broaden both the sparse-baseline set and the eligible pool. Post-training baselines include one-shot MP and Staged MP when the latter is present. These post-training baselines, small dense baselines, and dense references are reported separately and are not selection eligible. Protocol-level candidate accounting is given in the Seed and Candidate-Pool Details subsection.

All candidates are selected using held-out validation metrics; we recompute all metrics on the test split for the final tables. Selection groups without an in-band candidate are flagged explicitly, and final tables report the resulting test accuracy gap together with the behavioral coordinates.

\paragraph{Large-pool sparse baselines.}
The large-pool CIFAR-10 ResNet-18 protocol at 50\% sparsity expands the selection-eligible pool with multiple candidates from the LT baseline, LT reinitialization, global random sparse, layerwise random sparse, IMP, and SNIP methods \citep{han2015learning,lee2019snip}. The candidate composition is listed in the Seed and Candidate-Pool Details subsection below. The evaluation framework also supports GraSP and data-free SynFlow candidate types for pruning at initialization \citep{wang2020picking,tanaka2020pruning}, as well as RigL-style dynamic sparse training candidates \citep{evci2020rigging}. GraSP and SynFlow are not part of the aggregate comparisons reported here.

We measure the 200-epoch dynamic sparse training baseline in the CIFAR-10 ResNet-18 RigL protocol at 80\% sparsity, using the same sparse budget as the LT candidates. The aggregate 200-epoch comparisons reported here focus on the measured RigL baseline.

The RigL procedure starts from a global random sparse mask, periodically drops low-magnitude active weights, regrows inactive weights with large instantaneous gradients, and preserves the exact global active weight count. Protocol variants that change pruning fine-tuning learning rates or add the Staged MP baseline are tracked as separate protocol groups.

\paragraph{Selection-eligibility pool variants.}
The restricted-pool analysis reruns validation selection with ticket-family and LT-only eligibility criteria. The analysis quantifies behavior-aware selection for the operationally defined ticket-family or LT-only candidate sets. The ticket-family pool contains candidates from the LT baseline, LT reinitialization, IMP, and SNIP methods. The LT-only variant limits reselection to LT-baseline candidates. We report these reselection results alongside the primary all-sparse analysis. These post-hoc reselection outputs are derived rows and are not added to the primary artifact's 846-row accounting.

The result-quality summary reports, for every protocol--method group, the fraction of evaluations within the $\pm1.5$~pp test recovery band, the maximum absolute accuracy gap, and an indicator that all evaluations in that group satisfy the band. A non-strict group with an absolute mean accuracy gap of at most 3~pp is labeled near-recovery. These summaries distinguish strict accuracy-matched replacement evidence, near-recovery, and recovery-stress regimes.

Separate protocol keys specify the learning rates for dense-reference training, sparse-candidate training, post-pruning fine-tuning, and IMP. This protocol keying keeps stronger pruning-baseline sweeps in protocol groups keyed by the group-specific training and fine-tuning recipes.

\subsubsection{Seed and Candidate-Pool Details}

Unless otherwise stated, candidate counts below are per matched dense seed--sparsity cell, whereas table rows aggregate measured models over the dense seeds specified for each protocol. Accuracy-only and BP-LT rows are derived by selecting an already measured candidate for each dense reference; the derived rows add one reported row per selection group but no additional training run.

\paragraph{Model fields and reseeding.}
Each measured model entry records the dataset, architecture, sparsity, dense seed, candidate training seed, mask seed, and candidate source; a derived selection row additionally records the selected source. We use these fields together with the protocol keys defined above to assign training runs, behavioral-distance fits, and selection problems to the appropriate groups. The dense seed identifies the matched reference, the candidate training seed controls stochastic candidate training, and the mask seed records the seed used for mask construction when applicable.

For entries without a measured diagnostic, the corresponding field is left blank; optional identifiers such as mask seeds are retained when applicable. Evaluation transforms for validation selection are deterministic. For protocols that define disjoint auxiliary selection and evaluation splits, those split assignments are fixed explicitly.

\paragraph{Default sparse-candidate pools.}
For CIFAR-10 ResNet-18 and CIFAR-100 ResNet-18, the default selection-eligible pool contains six sparse candidates per dense seed--sparsity cell: one LT-baseline candidate trained from the original random initialization with the LT mask, one candidate trained with the same mask after random reinitialization, two candidates with global random sparse masks, and two candidates with layerwise random sparse masks that have the same per-layer sparsity as the corresponding LT mask. For CIFAR-10 ResNet-34 and CIFAR-10 WideResNet-28-2 \citep{zagoruyko2016wide}, the pool contains four candidates, one from each of those four method types. Separately reported one-shot MP and small dense rows are not included in these pool counts. When the restricted-pool analysis is limited to LT-baseline candidates, improvements depend on variation across the measured LT-candidate training seeds.

\paragraph{Large-pool and RigL protocols.}
At 50\% sparsity, the large-pool CIFAR-10 ResNet-18 evaluation comprises three dense seeds and 15 selection-eligible candidates per dense seed: three LT-baseline training seeds, two LT reinitializations, four candidates with global random sparse masks, four candidates with layerwise random sparse masks, one IMP candidate, and one SNIP candidate. Accordingly, Appendix Table~\ref{tab:large-pool} aggregates nine, six, twelve, twelve, three, and three measured candidates for these methods, respectively, along with three dense-reference runs. The three separately reported One-shot MP rows are not part of the 15-candidate eligible pool, and each selection rule contributes three derived rows.

At 80\% sparsity, the large-pool evaluation also includes three dense seeds. For each dense seed, the selection-eligible pool contains one LT-baseline candidate, one LT reinitialization, one candidate with a global random sparse mask, one candidate with a layerwise random sparse mask, one SNIP candidate, and one two-round IMP candidate with 20 fine-tuning epochs per round. Table~\ref{tab:large-pool-80} therefore aggregates three measured models per method and three derived rows per selection rule. One-shot MP and the four-stage Staged MP baseline are reported separately and are not selection eligible.

For the CIFAR-10 ResNet-18 RigL protocol at 80\% sparsity, we retain the same three dense seeds. RigL performs 132 topology updates per seed at the specified update interval and preserves the exact global active weight count. In the companion protocol with a lower post-pruning fine-tuning learning rate, LT and RigL are selection eligible, whereas one-shot MP and Staged MP are reported separately. Appendix Table~\ref{tab:rigl-80} aggregates three measured models for each method and three derived rows for each selection rule.

\paragraph{Replication counts.}
The primary CIFAR-10 ResNet-18 evaluation covers four sparsity levels (50\%, 80\%, 90\%, and 95\%) for five dense seeds. CIFAR-100 ResNet-18 at the 50\% and 80\% sparsity levels, CIFAR-10 ResNet-34 at 80\% sparsity, and CIFAR-10 WideResNet-28-2 at 80\% sparsity are evaluated for three dense seeds. The ConvNeXt-Tiny \citep{liu2022convnet} and ViT-Tiny \citep{dosovitskiy2021image} Imagenette protocols at 50\% sparsity are evaluated with three dense seeds and, for each dense seed, two training seeds for LT-baseline candidates; the LT entries therefore each summarize six candidates. The Flowers-102 \citep{nilsback2008automated} recovery-stress protocol at 80\% sparsity, the Flowers-102 protocol at 50\% sparsity, and the FGVC-Aircraft \citep{maji2013fine} protocol at 50\% sparsity are likewise evaluated with three dense seeds and two LT training seeds per dense seed. The LT entries for these three protocols therefore each summarize six candidates, whereas the dense, accuracy-only, BP-LT, and one-shot MP entries each summarize three. In both 50\% transfer protocols, the LT, accuracy-only, BP-LT, and one-shot MP protocol--method groups satisfy the study-defined accuracy-matching criterion; the random-mask controls do not.

\subsubsection{Behavioral Evaluation Coverage}

\paragraph{Corruption protocol coverage.}
Each corruption-accuracy evaluation records the evaluated corruption source explicitly. The recorded source distinguishes benchmark corruption data from synthetic Gaussian-noise stress tests; the benchmark data come from the CIFAR corruption family or directory-structured corruption datasets. In the standard CIFAR-10 rows with CIFAR-10-C coverage, we use Gaussian noise, shot noise, brightness, and contrast corruptions at severities 1, 3, and 5. For the large-pool CIFAR-10 protocols, we evaluate all 15 CIFAR-10-C corruption families at the same severities. For CIFAR-100, $\RobAcc$ denotes corruption accuracy under the Gaussian-noise stress test used in those rows.

\paragraph{Policy-coordinate availability.}
The $\mathrm{ReviewRate}_{\tau_D}$, $\mathrm{AutoErr}_{\tau_D}$, and $\mathrm{OODAccept}_{\tau_D}$ coordinates are available for the Imagenette ConvNeXt and ViT protocols, the fine-grained transfer protocols, and the large-pool CIFAR-10 protocols. The $\mathrm{AcceptFlip}_{\tau_D}$ and $\mathrm{PredDisagree}$ coordinates are available for the ConvNeXt protocol at 50\% sparsity, the ViT protocol at 50\% sparsity, the Flowers-102 transfer protocol at 50\% sparsity, the FGVC-Aircraft transfer protocol, and the large-pool CIFAR-10 protocols.

A coordinate enters the standardized distance when the dense reference value is finite. The coordinate must also be finite for at least one distance-fit row in the same protocol group. Coordinates not measured at the protocol level are inactive for that protocol. If a row is missing an otherwise active coordinate, we impute that coordinate with the protocol distance-fit mean before standardization.

\subsubsection{Statistical Summaries}
We report means and standard deviations across replicated evaluations, emphasize paired effect sizes and replication across sparsities and architectures, and include sensitivity analyses that apply rank distance and shrinkage-based Mahalanobis distance to the same behavioral-coordinate comparisons used by the diagonal distance. We treat these summaries as replicated descriptive evidence, not as formal statistical equivalence or non-inferiority tests. An operational release analysis should attach uncertainty intervals to predeclared coordinate-specific tolerances.

\subsubsection{Protocol Accounting and Compute}

Experiments were run on one Aurora compute node with six graphics processing units (GPUs), specifically Intel Data Center GPU Max 1550 devices, each comprising two independently addressable GPU tiles. The node therefore provides 12 tiles in total. We used PyTorch's \texttt{torch.xpu} backend, and each training process was assigned to one tile via \texttt{ZE\_AFFINITY\_MASK}.

The aggregate artifact contains 846 result rows from 56 experiment groups, comprising 624 non-selection rows and 222 derived validation-selection rows. A selection row carries the metrics of the selected candidate, so 846 is not the number of independent training runs. Method identity and protocol fields that encode choices affecting training or behavioral evaluation index each row.

\subsection{Supplemental CIFAR-10 Evidence}

\subsubsection{CIFAR-10 ResNet-18 Trade-Off Plot}

\begin{figure}[t!]
	\centering
	\includegraphics[width=0.96\linewidth]{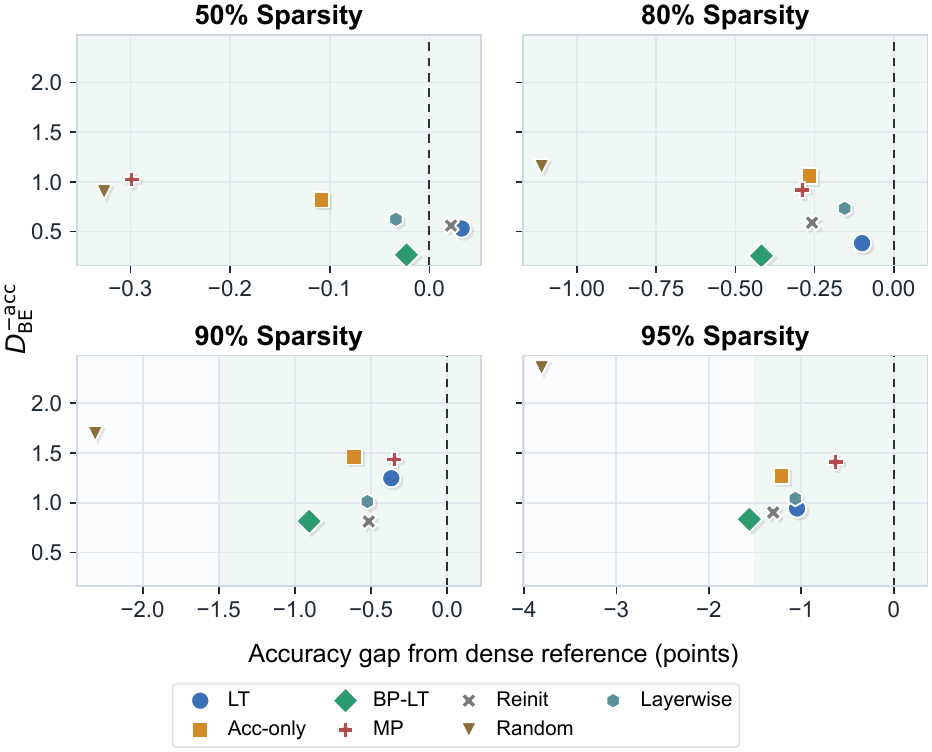}
	\caption{CIFAR-10 ResNet-18 accuracy gap and clean-accuracy-excluded behavioral-compatibility distance. Markers show method means over evaluated models with small display offsets to separate overlapping methods; the vertical reference line marks the dense reference's zero accuracy gap, and the shaded region marks the $\pm1.5$~pp test recovery band.}
	\label{fig:cifar10-tradeoff}
\end{figure}

Appendix Figure~\ref{fig:cifar10-tradeoff} shows the same results in a trade-off plot. The 50\% and 80\% sparsity levels are the cleanest replacement regimes. Several sparse methods are close to zero accuracy gap, yet the corresponding behavioral distances remain visibly above zero. The 90\% and 95\% sparsity levels show the high-sparsity regime where accuracy recovery starts to degrade.

\subsubsection{Per-Metric Behavioral Decomposition}

Aggregate distances are useful for selection, but the distance summaries can hide which behavioral coordinates move. Appendix Table~\ref{tab:metric-deltas} reports signed test-set differences from the dense reference for representative settings. The sign convention follows each coordinate's raw orientation. For example, lower NLL and lower $\IDFAR$ are improvements, whereas a positive representation distance indicates that the sparse candidate's representation differs from that of the dense reference.

The key point is that accuracy-matched LT baselines move several deployment-relevant behavioral coordinates at once. For CIFAR-10 ResNet-18 at 50\% sparsity, the LT baseline improves OOD AUROC and corruption accuracy slightly but changes representation geometry and $\IDFAR$. At 80\% sparsity, the LT baseline has similar accuracy but lower corruption accuracy. The WideResNet results show larger raw shifts because sparse recovery is harder.

\begin{table*}[t!]
	\centering
	\scriptsize
	\begin{tabular}{ll|rrrrrrr}
		\toprule
		Setting                             & Method                                     & \multicolumn{1}{|c}{$\Delta\ECE$} &
		\multicolumn{1}{c}{$\Delta\NLL$}    & \multicolumn{1}{c}{$\Delta\mathrm{OOD}$}   &
		\multicolumn{1}{c}{$\Delta\IDFAR$}  & \multicolumn{1}{c}{$\Delta\RepDist$}       &
		\multicolumn{1}{c}{$\Delta\RobAcc$} & \multicolumn{1}{c}{$\Delta\mathrm{Worst}$}                                                                                          \\
		\cmidrule(lr){3-9}
		                                    &                                            & \multicolumn{1}{|c}{(pp)}         &           & \multicolumn{1}{c}{AUROC} &
		\multicolumn{1}{c}{(pp)}            &                                            & \multicolumn{1}{c}{(pp)}          &
		\multicolumn{1}{c}{(pp)}                                                                                                                                                  \\
		\midrule
		ResNet-18 50\%                      & LT                                         & $-0.30$                           & $-0.0126$ & $+0.0102$                 & $-9.03$  &
		$+0.0343$                           & $+0.16$                                    & $+0.22$                                                                                \\
		                                    & BP-LT                                      & $-0.17$                           & $-0.0061$ & $+0.0013$                 & $-5.10$  &
		$+0.0366$                           & $-0.58$                                    & $+0.14$                                                                                \\
		\addlinespace[1pt]
		ResNet-18 80\%                      & LT                                         & $-0.01$                           & $+0.0040$ & $+0.0027$                 & $-6.59$  &
		$+0.0441$                           & $-1.57$                                    & $+0.12$                                                                                \\
		                                    & BP-LT                                      & $+0.06$                           & $+0.0114$ & $-0.0161$                 & $+3.09$  &
		$+0.0470$                           & $-1.94$                                    & $-0.22$                                                                                \\
		\addlinespace[1pt]
		WideResNet-28-2 80\%                & LT                                         & $+0.71$                           & $+0.0551$ & $-0.0535$                 & $+17.91$ &
		$+0.0950$                           & $-3.54$                                    & $-1.30$                                                                                \\
		                                    & BP-LT                                      & $+0.65$                           & $+0.0535$ & $-0.0243$                 & $+6.50$  &
		$+0.0915$                           & $-2.86$                                    & $-2.07$                                                                                \\
		\bottomrule
	\end{tabular}
	\caption{Per-metric signed candidate-minus-reference differences. Lower ECE, NLL, and $\IDFAR$ are better, whereas higher OOD AUROC, $\RobAcc$, and $\mathrm{WorstClassAcc}$ are better. A positive $\RepDist$ means that the candidate is farther from the dense reference in representation space. Worst denotes $\mathrm{WorstClassAcc}$.}
	\label{tab:metric-deltas}
\end{table*}

\subsubsection{Controls at 80\% and 95\% Sparsity Levels}

Appendix Table~\ref{tab:controls} gives additional CIFAR-10 ResNet-18 controls. At 80\%, one-shot MP fine-tuning reaches $94.26\%$ accuracy and a distance of $0.821$, LT reinitialization reaches a distance of $0.825$, and candidates with layerwise random sparse masks have a mean distance of $0.818$. These controls are all accuracy-competitive but behaviorally distinct from the dense reference on the measured audit coordinates. At 95\%, one-shot MP fine-tuning gives the best clean accuracy among sparse controls, but the behavioral distance of one-shot MP is larger than the distances of the LT baseline and BP-LT. This 95\% control result illustrates why we separate accuracy matching from behavioral matching.

\begin{table}[t!]
	\centering
	\scriptsize
	\begin{tabular}{cl|rrr}
		\toprule
		Sparsity (\%)                         & Method           & \multicolumn{1}{|c}{$\Acc$ (\%)} &
		\multicolumn{1}{c}{$\Delta\Acc$ (pp)} &
		\multicolumn{1}{c}{$\DBEacc$}                                                                                   \\
		\midrule
		\multirow{4}{*}{80}                   & One-shot MP      & $94.26$                          & $-0.36$ & $0.821$ \\
		                                      & LT reinit        & $94.35$                          & $-0.27$ & $0.825$ \\
		                                      & Layerwise random & $94.39$                          & $-0.23$ & $0.818$ \\
		                                      & Small dense      & $88.05$                          & $-6.57$ & $2.851$ \\
		\addlinespace[1pt]
		\multirow{4}{*}{95}                   & One-shot MP      & $93.76$                          & $-0.86$ & $1.313$ \\
		                                      & LT reinit        & $93.28$                          & $-1.34$ & $1.137$ \\
		                                      & Layerwise random & $93.30$                          & $-1.32$ & $1.126$ \\
		                                      & Small dense      & $88.05$                          & $-6.57$ & $2.801$ \\
		\bottomrule
	\end{tabular}
	\caption{CIFAR-10 ResNet-18 controls.}
	\label{tab:controls}
\end{table}

\subsubsection{Expanded 50\% Pool with IMP and SNIP}

The primary CIFAR-10 evaluation in Table~\ref{tab:cifar10-sparsity} is based on 200-epoch training with a compact sparse selection-eligible pool. We next test CIFAR-10 ResNet-18 at 50\% sparsity with a broader sparse set and auxiliary evaluation suite. Candidate composition and auxiliary-source details are in the Experimental Protocol and Reproducibility subsection above.

Appendix Table~\ref{tab:large-pool} shows that the larger selection-eligible pool makes clean-accuracy recovery easier, not harder. The dense reference reaches $94.55\%$ accuracy. The LT-baseline candidates average $94.70\%$. LT reinitialization, IMP, SNIP, global random sparse, layerwise random sparse, and one-shot MP baselines all remain within about 0.16~pp of the dense reference. Together, these groups provide study-band-matched evidence of measurable behavioral deviation. The~LT~baseline has $\DBEacc=0.952$, prediction disagreement $4.5\%$, and $\mathrm{AcceptFlip}_{\tau_D}=8.3\%$ for decisions made using the dense threshold. BP-LT reduces the behavioral distance to $0.832$ while preserving the dense reference's accuracy. The BP-LT selections draw from LT reinitialization, global random sparse, and SNIP candidates, so this row evaluates behavior-aware sparse selection. The LT rows isolate the behavioral incompatibility of LTs.

\begin{table}[t!]
	\centering
	\scriptsize
	\resizebox{\linewidth}{!}{%
		\begin{tabular}{l|rrrrrr}
			\toprule
			Method                        & \multicolumn{1}{|c}{$\Acc$}   & \multicolumn{1}{c}{$\Delta\Acc$} &
			\multicolumn{1}{c}{$\DBEacc$} & \multicolumn{1}{c}{OOD}       &
			\multicolumn{1}{c}{Flip}      & \multicolumn{1}{c}{$\RobAcc$}                                                                             \\
			                              & \multicolumn{1}{|c}{(\%)}     & \multicolumn{1}{c}{(pp)}         &         &
			\multicolumn{1}{c}{AUROC}     & \multicolumn{1}{c}{(\%)}      &
			\multicolumn{1}{c}{(\%)}                                                                                                                  \\
			\midrule
			Dense                         & $94.55\pm0.18$                & $0.00$                           & $0.000$ & $0.8688$ & $0.00$  & $71.75$ \\
			\addlinespace[1pt]
			LT                            & $94.70\pm0.16$                & $+0.16$                          & $0.952$ & $0.8630$ & $8.29$  & $71.85$ \\
			LT reinit                     & $94.62\pm0.12$                & $+0.07$                          & $0.989$ & $0.8648$ & $8.05$  & $71.83$ \\
			IMP                           & $94.50\pm0.23$                & $-0.05$                          & $1.096$ & $0.8599$ & $8.40$  & $72.11$ \\
			SNIP                          & $94.50\pm0.07$                & $-0.05$                          & $1.074$ & $0.8538$ & $8.12$  & $71.55$ \\
			Random sparse                 & $94.41\pm0.21$                & $-0.14$                          & $0.957$ & $0.8586$ & $8.49$  & $71.52$ \\
			Layerwise random              & $94.62\pm0.11$                & $+0.08$                          & $1.027$ & $0.8574$ & $8.33$  & $72.13$ \\
			One-shot MP                   & $94.52\pm0.06$                & $-0.03$                          & $1.054$ & $0.8542$ & $7.22$  & $71.73$ \\
			\addlinespace[2pt]
			Accuracy-only                 & $94.64\pm0.04$                & $+0.09$                          & $0.958$ & $0.8650$ & $8.13$  & $71.95$ \\
			BP-LT                         & $94.50\pm0.05$                & $-0.04$                          & $0.832$ & $0.8667$ & $8.29$  & $71.47$ \\
			\addlinespace[1pt]
			Small dense                   & $88.17\pm0.19$                & $-6.38$                          & $3.925$ & $0.7799$ & $21.90$ & $60.50$ \\
			\bottomrule
		\end{tabular}
	}
	\caption{Large-pool CIFAR-10 ResNet-18 at 50\% sparsity.}
	\label{tab:large-pool}
\end{table}

\subsubsection{RigL and Staged MP with Lower Fine-Tuning Learning Rate}

A companion CIFAR-10 ResNet-18 protocol at 80\% sparsity strengthens the dynamic sparse training and post-training pruning baselines with one LT-baseline candidate, one RigL-style dynamic sparse training candidate, a one-shot MP baseline, and a Staged MP baseline. This protocol keeps the sparse-candidate learning rate at $1\times10^{-1}$ and sets the post-pruning fine-tuning learning rate to $1\times10^{-2}$; the learning rates for sparse training and post-pruning fine-tuning define separate protocol keys. The LT and RigL candidates are trained with the same 200-epoch sparse training budget. RigL follows the specified update interval and preserves the exact global active weight count.

Appendix Table~\ref{tab:rigl-80} shows that alternative baseline hyperparameters change the post-training pruning picture, while the replacement conclusion remains unchanged. Staged MP now recovers the dense reference's accuracy, reaching $94.58\%$ compared with $94.66\%$ for the reference, and is the sparse baseline closest in behavior to the dense reference in this protocol, with $\DBEacc=1.577$, $\mathrm{PredDisagree}=1.6\%$, and $\mathrm{AcceptFlip}_{\tau_D}=3.2\%$. RigL also recovers clean accuracy at $94.50\%$ but shows a larger behavioral distance of $2.021$, $\mathrm{PredDisagree}=4.8\%$, and $\mathrm{AcceptFlip}_{\tau_D}=8.9\%$. The LT baseline is similarly accuracy-matched but farther from the dense reference's behavior in this protocol, with $\DBEacc=2.407$. Thus, even a candidate produced by strong dynamic sparse training remains behaviorally distinguishable from the dense reference after clean accuracy recovers.

\begin{table}[t!]
	\centering
	\scriptsize
	\resizebox{\linewidth}{!}{%
		\begin{tabular}{l|rrrrrrr}
			\toprule
			Method                        & \multicolumn{1}{|c}{$\Acc$} & \multicolumn{1}{c}{$\Delta\Acc$} &
			\multicolumn{1}{c}{$\DBEacc$} & \multicolumn{1}{c}{OOD}     &
			\multicolumn{1}{c}{Flip}      & \multicolumn{1}{c}{PredDis} &
			\multicolumn{1}{c}{$\RobAcc$}                                                                                                                   \\
			                              & \multicolumn{1}{|c}{(\%)}   & \multicolumn{1}{c}{(pp)}         &         &
			\multicolumn{1}{c}{AUROC}     & \multicolumn{1}{c}{(\%)}    &
			\multicolumn{1}{c}{(\%)}      & \multicolumn{1}{c}{(\%)}                                                                                        \\
			\midrule
			Dense                         & $94.66\pm0.07$              & $0.00$                           & $0.000$ & $0.8684$ & $0.00$ & $0.00$ & $72.24$ \\
			\addlinespace[1pt]
			LT                            & $94.50\pm0.17$              & $-0.16$                          & $2.407$ & $0.8599$ & $8.54$ & $4.44$ & $71.32$ \\
			One-shot MP                   & $94.52\pm0.20$              & $-0.14$                          & $1.736$ & $0.8662$ & $3.68$ & $1.71$ & $71.05$ \\
			Staged MP                     & $94.58\pm0.15$              & $-0.07$                          & $1.577$ & $0.8663$ & $3.23$ & $1.58$ & $71.56$ \\
			RigL                          & $94.50\pm0.07$              & $-0.16$                          & $2.021$ & $0.8614$ & $8.93$ & $4.83$ & $71.28$ \\
			\addlinespace[2pt]
			Accuracy-only                 & $94.46\pm0.09$              & $-0.19$                          & $2.045$ & $0.8623$ & $8.80$ & $4.68$ & $71.42$ \\
			BP-LT                         & $94.50\pm0.07$              & $-0.16$                          & $2.021$ & $0.8614$ & $8.93$ & $4.83$ & $71.28$ \\
			\bottomrule
		\end{tabular}
	}
	\caption{CIFAR-10 ResNet-18 at 80\% sparsity with RigL and Staged MP under the companion protocol with a lower post-pruning fine-tuning learning rate. LT and RigL candidates are trained for 200 epochs under the sparse-training recipe; pruning baselines follow the corresponding protocol-specific fine-tuning schedules. PredDis denotes $\mathrm{PredDisagree}$.}
	\label{tab:rigl-80}
\end{table}

\subsection{Replications and Transfer}

This subsection gives the full tables and per-setting interpretation for the replication and transfer results summarized in the main text. Appendix Table~\ref{tab:replications} reports full CIFAR-100, ResNet-34, and WideResNet replication results, and detailed interpretation of the pretrained settings is given below. The replications reproduce the pattern of accuracy recovery with behavioral gaps across changes in datasets and architectures. Appendix Table~\ref{tab:imagenette50} provides the pretrained replication. ConvNeXt-Tiny and ViT-Tiny both recover the dense reference's accuracy at 50\% sparsity while retaining nonzero behavioral distance and dense-threshold acceptance flips.

\subsubsection{CIFAR-100 and Architecture Replications}

Appendix Table~\ref{tab:replications} reports the harder CIFAR-100, deeper ResNet-34, and WideResNet settings. CIFAR-100 at 50\% sparsity is the most important additional accuracy-matched result. The dense reference reaches $75.30\%$, the LT baseline reaches $74.66\%$, and BP-LT reaches $74.95\%$. The behavior-aware selection remains closer in behavior than the LT baseline.

At 80\% sparsity, CIFAR-100 still shows the behavioral pattern as recovery becomes harder. The LT baseline and BP-LT are 1.55 and 1.50~pp, respectively, below the dense reference on average. Some individual evaluation gaps exceed the prespecified $\pm1.5$~pp test recovery band, but the mean gaps remain within the near-recovery threshold of 3~pp; these protocol--method groups are therefore classified as near-recovery.

CIFAR-10 ResNet-34 at 80\% sparsity replicates the main phenomenon with a stronger dense reference. The LT baseline is within 0.07~pp of the reference but has a behavioral distance of $1.029$. BP-LT lowers the behavioral distance to $0.574$ while remaining within 0.31~pp of the reference.

CIFAR-10 WideResNet-28-2 is a harder architecture stress case. The dense reference reaches $94.18\%$, while the 80\% LT baseline reaches $93.01\%$. This WideResNet LT baseline remains within the prespecified $\pm1.5$~pp test recovery band with a larger 1.17~pp gap. The LT baseline is behaviorally distinct from the corresponding dense reference on the measured audit panel, with $\DBEacc=1.149$. BP-LT lowers this distance to $1.014$. This result shows why architecture diversity, band status, and the actual accuracy gap should be reported together: Recovery can be marginal even when the test recovery band used in this paper is satisfied.

\begin{table}[t!]
	\centering
	\scriptsize
	\begin{tabular}{l|rrrr}
		\toprule
		Method                                & \multicolumn{1}{|c}{$\Acc$ (\%)} &
		\multicolumn{1}{c}{$\Delta\Acc$ (pp)} &
		\multicolumn{1}{c}{$\DBEacc$}         &
		\multicolumn{1}{c}{$\RobAcc$ (\%)}                                                                     \\
		\midrule
		\multicolumn{5}{l}{\textbf{CIFAR-100 ResNet-18, 50\% sparsity}}                                        \\
		\cmidrule(lr){1-5}
		Dense                                 & $75.30$                          & $0.00$  & $0.000$ & $43.50$ \\
		LT                                    & $74.66$                          & $-0.64$ & $0.768$ & $42.10$ \\
		BP-LT                                 & $74.95$                          & $-0.35$ & $0.548$ & $42.29$ \\
		\addlinespace[1pt]
		\multicolumn{5}{l}{\textbf{CIFAR-100 ResNet-18, 80\% sparsity}}                                        \\
		\cmidrule(lr){1-5}
		Dense                                 & $75.30$                          & $0.00$  & $0.000$ & $43.50$ \\
		LT                                    & $73.75$                          & $-1.55$ & $1.005$ & $40.06$ \\
		BP-LT                                 & $73.80$                          & $-1.50$ & $0.894$ & $40.51$ \\
		One-shot MP                           & $74.55$                          & $-0.75$ & $1.243$ & $42.09$ \\
		\addlinespace[1pt]
		\multicolumn{5}{l}{\textbf{CIFAR-10 ResNet-34, 80\% sparsity}}                                         \\
		\cmidrule(lr){1-5}
		Dense                                 & $94.86$                          & $0.00$  & $0.000$ & $67.75$ \\
		LT                                    & $94.79$                          & $-0.07$ & $1.029$ & $68.29$ \\
		Accuracy-only                         & $94.70$                          & $-0.15$ & $0.723$ & $67.50$ \\
		BP-LT                                 & $94.54$                          & $-0.31$ & $0.574$ & $66.56$ \\
		One-shot MP                           & $94.74$                          & $-0.11$ & $0.483$ & $68.12$ \\
		\addlinespace[1pt]
		\multicolumn{5}{l}{\textbf{CIFAR-10 WideResNet-28-2, 80\% sparsity}}                                   \\
		\cmidrule(lr){1-5}
		Dense                                 & $94.18$                          & $0.00$  & $0.000$ & $65.21$ \\
		LT                                    & $93.01$                          & $-1.17$ & $1.149$ & $61.67$ \\
		Accuracy-only                         & $92.94$                          & $-1.24$ & $1.131$ & $61.45$ \\
		BP-LT                                 & $93.04$                          & $-1.14$ & $1.014$ & $62.34$ \\
		One-shot MP                           & $93.53$                          & $-0.65$ & $1.071$ & $64.12$ \\
		\bottomrule
	\end{tabular}
	\caption{Results for additional dataset and architecture settings. Each entry is the mean over evaluated models.}
	\label{tab:replications}
\end{table}

\subsubsection{Detailed ConvNeXt and ViT Results}

\begin{table}[t!]
	\centering
	\scriptsize
	\resizebox{\linewidth}{!}{%
		\begin{tabular}{l|rrrrrr}
			\toprule
			Method                        & \multicolumn{1}{|c}{$\Acc$}   & \multicolumn{1}{c}{$\Delta\Acc$} &
			\multicolumn{1}{c}{$\DBEacc$} & \multicolumn{1}{c}{OOD}       &
			\multicolumn{1}{c}{Flip}      & \multicolumn{1}{c}{$\RobAcc$}                                                                            \\
			                              & \multicolumn{1}{|c}{(\%)}     & \multicolumn{1}{c}{(pp)}         &         &
			\multicolumn{1}{c}{AUROC}     & \multicolumn{1}{c}{(\%)}      &
			\multicolumn{1}{c}{(\%)}                                                                                                                 \\
			\midrule
			\multicolumn{7}{l}{\textbf{ConvNeXt-Tiny}}                                                                                               \\
			\cmidrule(lr){1-7}
			Dense                         & $97.49\pm0.14$                & $0.00$                           & $0.000$ & $0.843$ & $0.00$  & $96.60$ \\
			\addlinespace[1pt]
			LT                            & $97.94\pm0.11$                & $+0.45$                          & $1.205$ & $0.829$ & $9.69$  & $97.28$ \\
			Accuracy-only                 & $97.98\pm0.15$                & $+0.49$                          & $1.436$ & $0.785$ & $9.75$  & $97.29$ \\
			BP-LT                         & $97.90\pm0.06$                & $+0.42$                          & $0.974$ & $0.873$ & $9.62$  & $97.27$ \\
			One-shot MP                   & $97.47\pm0.08$                & $-0.02$                          & $0.716$ & $0.895$ & $11.74$ & $96.52$ \\
			\addlinespace[1pt]
			Random sparse                 & $87.50\pm0.04$                & $-9.99$                          & $1.577$ & $0.758$ & $34.37$ & $84.32$ \\
			Layerwise random              & $81.69\pm0.92$                & $-15.80$                         & $2.236$ & $0.732$ & $49.26$ & $81.07$ \\
			\addlinespace[2pt]
			\multicolumn{7}{l}{\textbf{ViT-Tiny}}                                                                                                    \\
			\cmidrule(lr){1-7}
			Dense                         & $97.01\pm0.32$                & $0.00$                           & $0.000$ & $0.833$ & $0.00$  & $96.51$ \\
			\addlinespace[1pt]
			LT                            & $96.73\pm0.12$                & $-0.28$                          & $0.422$ & $0.861$ & $8.08$  & $95.82$ \\
			Accuracy-only                 & $96.75\pm0.16$                & $-0.26$                          & $0.432$ & $0.857$ & $8.21$  & $95.88$ \\
			BP-LT                         & $96.67\pm0.17$                & $-0.34$                          & $0.376$ & $0.854$ & $8.08$  & $95.66$ \\
			One-shot MP                   & $96.06\pm0.06$                & $-0.95$                          & $0.464$ & $0.776$ & $11.02$ & $95.03$ \\
			\addlinespace[1pt]
			Random sparse                 & $83.54\pm0.29$                & $-13.47$                         & $1.975$ & $0.768$ & $52.08$ & $81.93$ \\
			Layerwise random              & $81.24\pm0.37$                & $-15.77$                         & $2.130$ & $0.737$ & $53.35$ & $80.30$ \\
			\bottomrule
		\end{tabular}
	}
	\caption{Imagenette evaluations with pretrained ConvNeXt-Tiny and ViT-Tiny at 50\% sparsity on 224-pixel inputs. Auxiliary selection and evaluation splits are disjoint, and corruption accuracy is evaluated under the Imagenette Gaussian-noise protocol.}
	\label{tab:imagenette50}
\end{table}

\paragraph{ConvNeXt Recovery with Disjoint Auxiliary Splits.}

To evaluate behavioral compatibility with a strong pretrained dense reference, we evaluate an Imagenette ConvNeXt-Tiny protocol at 50\% sparsity using 224-pixel inputs and disjoint auxiliary selection and evaluation splits. The pretrained training recipe and candidate composition are described in the Experimental Protocol and Reproducibility subsection above.

The ConvNeXt rows in Appendix Table~\ref{tab:imagenette50} show the result. The dense reference reaches $97.49\%$ accuracy. The LT baseline at 50\% sparsity reaches $97.94\%$ mean test accuracy, slightly above the dense reference, yet the LT baseline's clean-accuracy-excluded behavioral-compatibility distance remains $1.205$. BP-LT selects a candidate that is closer under the validation criterion, reducing the distance to $0.974$ while maintaining $97.90\%$ accuracy. Accuracy-only selection chooses the candidate with the closest validation accuracy but is behaviorally farther from the dense reference, with a distance of $1.436$. One-shot MP fine-tuning almost exactly recovers the dense reference's accuracy and has a distance of $0.716$, showing that a conventional pruning baseline can be closer in behavior in this particular setting.

The global random sparse and layerwise random sparse controls show that active parameter count alone does not explain the ConvNeXt behavioral gap. Both controls are trained for the same 100 epochs but reach $87.50\%$ and $81.69\%$ test accuracy, respectively, with much larger $\mathrm{AcceptFlip}_{\tau_D}$ and $\mathrm{PredDisagree}$ rates.

The ConvNeXt rows in Appendix Table~\ref{tab:imagenette50} show that recovery stress does not explain the behavioral gap. Here the LT baseline exceeds the dense reference's clean accuracy, but replacing the dense reference still flips about 9.7\% of dense-threshold acceptance decisions and disagrees with the dense reference's predictions on about 2.5\% of test examples. The combination of higher clean accuracy with policy and prediction shifts is the operating regime addressed by our behavioral- compatibility audit.

\paragraph{ViT Recovery at 50\% Sparsity.}

The pretrained ViT-Tiny evaluation extends the recovery pattern to transformer backbones evaluated at 50\% sparsity using 224-pixel inputs and disjoint auxiliary selection and evaluation splits. The ViT rows in Appendix Table~\ref{tab:imagenette50} report the result. The dense reference reaches $(97.01\pm0.32)\%$, and the LT baseline reaches $(96.73\pm0.12)\%$, a 0.28~pp mean accuracy gap. Thus, 50\% sparsity constitutes a transformer-backed setting matched under the study-defined accuracy rule.

The recovered LT baseline for ViT-Tiny in Appendix Table~\ref{tab:imagenette50} is still not behaviorally interchangeable with the dense reference: $\DBEacc=0.422$, $\mathrm{AcceptFlip}_{\tau_D}$ is about $8.1\%$, and top-1 predictions disagree on about $2.9\%$. BP-LT selects a candidate closer in behavior, reducing the distance to $0.376$ while keeping accuracy at $96.67\%$. One-shot MP fine-tuning is also strong in clean accuracy at $96.06\%$ but has a larger behavioral distance and larger $\mathrm{AcceptFlip}_{\tau_D}$ rate. Global random sparse and layerwise random sparse controls achieve test accuracies of $83.54\%$ and $81.24\%$, respectively. The ViT result therefore adds a transformer-backed case matched under the study-defined accuracy rule. Once recovery occurs, clean accuracy still does not imply behavioral compatibility.

\subsubsection{Fine-Grained Transfer}

Appendix Table~\ref{tab:transfer} reports fine-grained transfer evaluations of pretrained ConvNeXt-Tiny models on 224-pixel inputs and identifies the protocol--method groups satisfying the study-defined accuracy-matching criterion at 50\% sparsity. At 80\% sparsity on Flowers-102, the LT baseline reaches $68.18\%$ accuracy compared with $98.24\%$ for the dense reference, marking a transfer recovery-stress point.

The Flowers-102 protocol at 50\% sparsity extends the non-CIFAR evidence beyond Imagenette. The LT, accuracy-only, BP-LT, and one-shot MP protocol--method groups of this protocol are matched under the study-defined accuracy rule. The dense ConvNeXt-Tiny reference reaches $(98.51\pm0.13)\%$, the LT baseline reaches $(98.24\pm0.46)\%$, and accuracy-only selection reaches $98.37\%$. Clean accuracy is now within 0.5~pp, while the behavioral gap remains. The LT baseline has $\DBEacc=0.739$ and $\mathrm{AcceptFlip}_{\tau_D}=7.28\%$, while BP-LT lowers the behavioral distance to $0.509$ at $98.23\%$ accuracy.

One-shot MP fine-tuning has a lower behavioral distance of $0.388$. One-shot MP remains within the same $\pm1.5$~pp test recovery band and has $\mathrm{AcceptFlip}_{\tau_D}=8.70\%$. Thus, a conventional pruning baseline can have a smaller aggregate behavioral distance while still changing the dense-threshold policy. Global random sparse and layerwise random sparse candidates reach only $57.56\%$ and $51.26\%$ accuracy.

FGVC-Aircraft at 50\% sparsity in Appendix Table~\ref{tab:transfer} gives a second transfer protocol in which the LT, accuracy-only, BP-LT, and one-shot MP groups are matched under the study-defined accuracy rule. The dense ConvNeXt-Tiny reference reaches $(85.18\pm0.18)\%$, the LT baseline reaches $(84.25\pm0.17)\%$, and accuracy-only selection reaches $84.29\%$. The sparse accuracy gap is about 1~pp. The LT baseline still differs from the dense reference's behavior. The LT baseline has $\DBEacc=0.658$, while BP-LT reduces the distance to $0.399$ at $84.19\%$ accuracy. Candidates with global random sparse and layerwise random sparse masks at the same sparsity reach only $39.05\%$ and $25.86\%$, respectively, so this transfer result also distinguishes candidates trained with the LT mask from those trained with generic sparse masks at the same active parameter count.

\begin{table}[t!]
	\centering
	\scriptsize
	\resizebox{\linewidth}{!}{%
		\begin{tabular}{l|rrrrr}
			\toprule
			Method                                & \multicolumn{1}{|c}{$\Acc$ (\%)} &
			\multicolumn{1}{c}{$\Delta\Acc$ (pp)} &
			\multicolumn{1}{c}{$\DBEacc$}         &
			\multicolumn{1}{c}{OOD AUROC}         & \multicolumn{1}{c}{Flip (\%)}                                              \\
			\midrule
			\multicolumn{6}{l}{\textbf{Flowers-102, 50\% sparsity}}                                                            \\
			\cmidrule(lr){1-6}
			Dense                                 & $98.51$                          & $0.00$   & $0.000$ & $0.9815$ & $0.00$  \\
			\addlinespace[1pt]
			LT                                    & $98.24$                          & $-0.27$  & $0.739$ & $0.9831$ & $7.28$  \\
			Accuracy-only                         & $98.37$                          & $-0.14$  & $0.761$ & $0.9808$ & $7.29$  \\
			BP-LT                                 & $98.23$                          & $-0.28$  & $0.509$ & $0.9855$ & $7.18$  \\
			One-shot MP                           & $97.60$                          & $-0.91$  & $0.388$ & $0.9819$ & $8.70$  \\
			\addlinespace[1pt]
			Random sparse                         & $57.56$                          & $-40.95$ & $1.853$ & $0.7009$ & $79.18$ \\
			Layerwise random                      & $51.26$                          & $-47.25$ & $2.153$ & $0.7278$ & $79.49$ \\
			\addlinespace[2pt]
			\multicolumn{6}{l}{\textbf{FGVC-Aircraft, 50\% sparsity}}                                                          \\
			\cmidrule(lr){1-6}
			Dense                                 & $85.18$                          & $0.00$   & $0.000$ & $0.9557$ & $0.00$  \\
			\addlinespace[1pt]
			LT                                    & $84.25$                          & $-0.93$  & $0.658$ & $0.9426$ & $7.91$  \\
			Accuracy-only                         & $84.29$                          & $-0.89$  & $0.670$ & $0.9354$ & $7.92$  \\
			BP-LT                                 & $84.19$                          & $-0.99$  & $0.399$ & $0.9400$ & $7.76$  \\
			One-shot MP                           & $84.09$                          & $-1.09$  & $0.472$ & $0.9572$ & $8.09$  \\
			\addlinespace[1pt]
			Random sparse                         & $39.05$                          & $-46.12$ & $1.587$ & $0.5472$ & $62.51$ \\
			Layerwise random                      & $25.86$                          & $-59.32$ & $1.808$ & $0.5688$ & $66.25$ \\
			\bottomrule
		\end{tabular}
	}
	\caption{Pretrained ConvNeXt-Tiny transfer evaluations with 224-pixel inputs. Sparse candidates are trained for 100 epochs, whereas one-shot MP follows a separate 20-epoch fine-tuning schedule. In both 50\% protocols, the LT, accuracy-only, BP-LT, and one-shot MP groups satisfy the study-defined accuracy-matching criterion; the random-mask controls are recovery-stress rows.}
	\label{tab:transfer}
\end{table}

\subsection{Selection and Distance Robustness}

\subsubsection{Selection-Source and Restricted-Pool Analysis}

\paragraph{Selection-source distribution.}
Across the 74 selection groups in the evaluation suite, accuracy-only selection chooses LT-baseline candidates in 35 groups, corresponding to $47.3\%$. BP-LT chooses LT-baseline candidates in 37 groups, corresponding to $50.0\%$. The remaining BP-LT selections come from LT reinitialization, global random sparse, layerwise random sparse, IMP, SNIP, or RigL candidates. These counts make the role of BP-LT explicit. In the all-sparse analysis, BP-LT performs behavior-aware validation selection within the evaluated sparse pool, while the LT rows provide the direct replacement audit for LTs.

Of the reported all-sparse BP-LT selections not drawn from LT-baseline candidates, 8 are LT-reinitialization candidates, 3 are RigL candidates, 1 is a SNIP candidate, and 25 are global random sparse or layerwise random sparse controls.

\paragraph{Restricted-pool analysis.}
A restricted-pool analysis applies the validation-time selection rules to measured candidates after restricting the eligible pool to either the ticket-family or LT-only pool. Under ticket-family eligibility, BP-LT selects LT-baseline candidates in 52 of the 74 groups. Ticket-family BP-LT selects LT-reinitialization candidates in 19 groups, IMP candidates in two groups, and a SNIP candidate in one group. These results show that the same qualitative conclusion holds under ticket-family eligibility.

For CIFAR-10 ResNet-18 at 50\% sparsity, ticket-family BP-LT lowers $\DBEacc$ from $0.648$ to $0.594$, compared with $0.446$ for all-sparse BP-LT. At 80\% sparsity, ticket-family BP-LT lowers the distance from $0.501$ to $0.488$, while all-sparse BP-LT lowers the LT baseline's distance to $0.437$. In the larger CIFAR-10 pool at 50\% sparsity, ticket-family BP-LT remains close to the all-sparse BP-LT result, with $0.854$ compared with $0.832$. In the large-pool protocol at 80\% sparsity, ticket-family BP-LT lowers the LT baseline's distance from $0.784$ to $0.728$, while all-sparse BP-LT lowers the LT baseline's distance to $0.701$.

In the LT-only pool, the achievable improvement is determined by variation among measured LT candidates. The analysis is reselection among already measured models, which isolates eligibility effects within the fixed measured pool.

\subsubsection{Distance Sensitivity}

Correlated coordinates can receive repeated weight in the diagonal summary; we therefore report coordinate-level signed results and evaluate rank- and covariance-aware distance variants here. The main distance is computed with diagonal standardization. Appendix Table~\ref{tab:sensitivity} shows that the conclusion is stable under alternative scalings. BP-LT is either competitive with the best method or the best overall under both rank distance and shrinkage-based Mahalanobis distance across all four CIFAR-10 ResNet-18 sparsities. The absolute values change, as expected, because rank distance suppresses outliers and Mahalanobis distance accounts for cross-coordinate covariance.

For both sensitivity analyses, we retain only those coordinates for which the dense-reference value is finite and at least one non-selection fitting row in the same protocol group has a finite value. Let $J\geq1$ denote the number of retained coordinates. For ranks and candidate-to-reference differences, missing evaluated-row values are imputed with the coordinate mean over all finite evaluated-row values in the group before forming the direction-signed vector $u(f)\in\mathbb{R}^{J}$. For the covariance fit, missing fitting-row values are instead imputed with the coordinate mean over the finite non-selection fitting-row values, giving $u^{\mathrm{fit}}(g)$. These vectors are separate from the standardized coordinates $Z_j$ used in the main distance. Let $r_j(f)$ be the percentile rank of $u_j(f)$ among evaluated models in the same protocol group, and let the Ledoit--Wolf (LW) shrinkage covariance, denoted by $\widehat{\Sigma}_{\mathrm{LW}}$, be estimated from the vectors $u^{\mathrm{fit}}(g)$ associated with non-selection rows \citep{ledoit2004well}. Here, non-selection rows are the sensitivity-analysis fitting rows after excluding rows produced by validation selection. Write $\widehat{\Sigma}_{\mathrm{LW}}^{\dagger}$ for the Moore--Penrose pseudoinverse of $\widehat{\Sigma}_{\mathrm{LW}}$, namely the precision matrix used by the implementation. The reported alternatives are
\begin{align*}
	D_{\mathrm{rank}}(f,f_D)
	 & \coloneqq
	\left(
	\frac{1}{J}\sum_{j=1}^{J}
	[r_j(f)-r_j(f_D)]^2
	\right)^{1/2},
\end{align*}
and
\begin{align*}
	D_{\mathrm{Mah}}(f,f_D)
	 & \coloneqq
	\left(
	\frac{1}{J}
	\delta(f)^{\top}
	\widehat{\Sigma}_{\mathrm{LW}}^{\dagger}
	\delta(f)
	\right)^{1/2}, \\
	\delta(f)
	 & \coloneqq
	u(f)-u(f_D).
\end{align*}

\begin{table}[t!]
	\centering
	\scriptsize
	\begin{tabular}{cl|rrr}
		\toprule
		Sparsity (\%)                           & Method        & \multicolumn{1}{|c}{$\DBEacc$} &
		\multicolumn{1}{c}{$D_{\mathrm{rank}}$} &
		\multicolumn{1}{c}{$D_{\mathrm{Mah}}$}                                                                       \\
		\midrule
		\multirow{3}{*}{50}                     & LT            & $0.648$                        & $0.412$ & $0.611$ \\
		                                        & Accuracy-only & $0.654$                        & $0.380$ & $0.591$ \\
		                                        & BP-LT         & $0.446$                        & $0.336$ & $0.437$ \\
		\addlinespace[1pt]
		\multirow{3}{*}{80}                     & LT            & $0.501$                        & $0.379$ & $0.413$ \\
		                                        & Accuracy-only & $0.890$                        & $0.432$ & $0.725$ \\
		                                        & BP-LT         & $0.437$                        & $0.403$ & $0.314$ \\
		\addlinespace[1pt]
		\multirow{3}{*}{90}                     & LT            & $1.364$                        & $0.503$ & $0.842$ \\
		                                        & Accuracy-only & $1.295$                        & $0.462$ & $0.779$ \\
		                                        & BP-LT         & $0.995$                        & $0.425$ & $0.517$ \\
		\addlinespace[1pt]
		\multirow{3}{*}{95}                     & LT            & $1.059$                        & $0.383$ & $0.743$ \\
		                                        & Accuracy-only & $1.099$                        & $0.386$ & $0.781$ \\
		                                        & BP-LT         & $1.014$                        & $0.371$ & $0.666$ \\
		\bottomrule
	\end{tabular}
	\caption{CIFAR-10 ResNet-18 distance sensitivity.}
	\label{tab:sensitivity}
\end{table}

\subsection{Additional Discussion}

\subsubsection{Main Findings}
The experiments support four conclusions. First, clean-accuracy matching can coexist with behavioral mismatch and lower performance under deployment-relevant shift. For CIFAR-10 ResNet-18 at 50\% sparsity, the LT baseline matches the dense reference's accuracy to within 0.01~pp but has $\DBEacc$ near 0.65. The same pattern persists at 80\% and appears again in CIFAR-10 ResNet-34. The signed $\RobAcc$ comparisons in Table~\ref{tab:cifar10-sparsity} and Appendix Tables~\ref{tab:replications} and~\ref{tab:imagenette50} additionally show directional losses for accuracy-matched LTs: 1.57~pp for CIFAR-10 ResNet-18 at 80\%, 1.40~pp for CIFAR-100 ResNet-18 at 50\%, 3.54~pp for CIFAR-10 WideResNet-28-2 at 80\%, and 0.69~pp for Imagenette ViT-Tiny at 50\%. The CIFAR-10 50\% and Imagenette ConvNeXt-Tiny LTs match or exceed their dense references' $\RobAcc$ values, showing that the direction varies by setting. Together, these results establish that clean recovery can conceal lower corruption performance and motivate signed, coordinate-level evaluation.

Second, BP-LT turns the behavioral audit into a practical validation-time selection rule. BP-LT lowers mean behavioral distance relative to the LT baseline and accuracy-only selection across the primary CIFAR-10 multi-sparsity evaluation, the larger CIFAR-10 candidate-pool protocols, and the pretrained Imagenette study-band-matched protocols while maintaining the accuracy constraint. Sensitivity analyses using rank distance and shrinkage-based Mahalanobis distance keep the same qualitative picture. Cases where conventional pruning baselines are closer in behavior further establish the audit as method-agnostic. The result is a method-agnostic operating-point compatibility audit paired with a validation-time sparse-candidate screening rule. Because BP-LT selects for behavioral closeness rather than directional utility, a smaller $\DBEacc$ should not be read as an improvement in every coordinate; deployment-specific utility requirements still require signed metric checks or explicit performance floors.

The coordinate-level results make this distinction concrete. At 50\% and 80\% sparsity for CIFAR-10 ResNet-18, BP-LT lowers mean $\DBEacc$ from $0.648$ to $0.446$ and from $0.501$ to $0.437$, respectively, while the BP-LT $\RobAcc$ is lower than that of the corresponding LT baseline. In the large-pool 80\% protocol, BP-LT lowers $\DBEacc$ from $0.784$ to $0.701$, but the mean BP-LT $\mathrm{AcceptFlip}_{\tau_D}$ is $8.54\%$, slightly above the LT baseline's $8.47\%$. Such outcomes are consistent with the objective: A reduction in the root-mean-square standardized distance can trade a larger deviation on one coordinate for smaller deviations on others, and validation-time selection need not preserve coordinate-wise ordering on the test distribution. Pareto dominance and worst-coordinate guarantees require additional constraints. Applications with asymmetric costs should combine the aggregate distance with coordinate-specific non-inferiority floors, application-weighted objectives, or Pareto screening.

The default all-sparse BP-LT analysis evaluates candidate screening within the measured pool. BP-LT selects an LT-baseline candidate in 37 of 74 selection groups; the remaining choices come from other eligible sparse sources. Ticket-family and LT-only analyses isolate progressively stricter eligibility, but the gains attainable under those pools are limited by variation in the measured pool. Thus, the all-sparse results show that behavior-aware selection can find a closer sparse replacement; improving LT construction is a separate training objective.

Third, the phenomenon is not explained by parameter count alone. The fixed-width small dense baselines are far from the dense reference in both accuracy and behavior, whereas sparse candidates with the same number of active parameters can achieve similar accuracy yet differ in behavior.

Fourth, results from harder settings characterize the transition between sparse recovery and behavioral replacement. Results for CIFAR-100 at 50\% sparsity support an accuracy-matched behavioral-compatibility analysis. Results at 80\% sparsity show the behavioral pattern as sparse recovery begins to degrade. The WideResNet-28-2 result adds a similar stress case. The LT baseline remains a strong classifier and stays within the test recovery band used in this paper, but the WideResNet accuracy gap is much closer to the tolerance edge than in the ResNet-18 settings at the 50\% and 80\% sparsity levels.

\subsubsection{Policy-Level Shift Interpretation}

Under this operating-point proxy, a flip rate of 8\% means that reusing $\tau_D$ after sparse replacement routes about 8 of every 100 in-distribution test inputs differently, which is approximately $0.08N$ out of $N$ processed inputs: Automatic acceptance becomes review, or review becomes automatic acceptance. The first direction can add review-queue demand and latency; the second bypasses a review that the dense policy would have required and can alter risk among automatically accepted inputs. Because $\mathrm{AcceptFlip}_{\tau_D}$ pools both directions, the acceptance-flip coordinate measures non-preservation of the routing policy rather than a net change in review rate or a signed utility change and is distinct from a top-1 label flip. The accompanying review-rate, automatically accepted error, and OOD acceptance coordinates help interpret the operational consequences. These experiments show routing non-preservation under the audited proxy policy. Translating the observed flips into reviewer outcomes, queueing cost, latency, or application-specific harm requires direction-specific flip rates and signed policy coordinates. Estimating a sparsity-specific causal effect additionally requires an independently retrained dense challenger to quantify excess churn beyond an ordinary model update. The present result establishes that accuracy recovery alone does not certify preservation of the audited action. The coexistence of these flips with small prediction-disagreement rates is the regime allowed by Theorem~\ref{thm:policy-nonidentifiability}. Threshold crossings need not be label changes. Theorem~\ref{thm:threshold-amplification} further gives the local first-order expansion under the stated path and density conditions.

\subsubsection{Theoretical Explanation of the Empirical Pattern}
Our theoretical results address two complementary questions: why the usual LT success criterion is insufficient for policy-compatible replacement and how confidence shifts can change deployment decisions. Theorem~\ref{thm:policy-nonidentifiability} shows that even when every top-1 prediction agrees, reusing the dense reference's fixed confidence threshold can change accept--review decisions for any fraction of inputs. Proposition~\ref{prop:threshold-flip-bound} gives a finite-perturbation flip bound, while Theorem~\ref{thm:threshold-amplification} makes the local policy-flip behavior quantitative. Along the stated local path, flip probability has a right-sided first-order expansion in the confidence perturbation, with $\kappa_{\tau_D}$ quantifying local sensitivity at the dense threshold. This local characterization complements the directly measured finite-model flip rates by explaining how confidence movement near the threshold can produce nonzero policy flips even when prediction disagreement is smaller. Together these results explain why clean-accuracy recovery alone leaves confidence-based deployment policies underdetermined.

This theory establishes a limitation of the certificate rather than a sparsity-dependent lower bound on behavioral distance. The results are mask-agnostic and leave open whether a sufficiently expressive sparse model can preserve the measured behavior. The empirical variation across LT recipes, conventional pruning baselines, and sparsity levels, together with the gains from BP-LT, instead suggests that part of the observed gap is mitigable. The residual gap may reflect candidate-pool coverage, objectives that do not train for reference compatibility, validation noise, or capacity limits at a given sparsity. Three questions remain open: whether a jointly compatible sparse candidate exists, whether a training procedure can find such a candidate, and whether validation data can select such a candidate.

\subsubsection{Sparse Baselines and Selection Evidence}
The large-pool CIFAR-10 protocols strengthen the first two conclusions with a broader baseline set. IMP, SNIP, global random sparse, and layerwise random sparse candidates recover the dense reference's accuracy at 50\% sparsity under the 200-epoch sparse training protocol. The separately scheduled one-shot MP baseline also recovers the reference's accuracy, giving a stronger sparse-baseline comparison.

At 80\% sparsity, the LT baseline, LT reinitialization, IMP, SNIP, global random sparse, layerwise random sparse, and one-shot MP protocol--method groups again satisfy the test recovery band and thus provide study-band-matched evidence relative to a $94.53\%$ dense reference. Even then, clean-accuracy-excluded behavioral distances persist, and dense-threshold $\mathrm{AcceptFlip}_{\tau_D}$ rates show that acceptance decisions continue to change.

Under the 80\% large-pool recipe, Staged MP falls in the recovery-stress stratum at $91.48\%$. Under the companion protocol with RigL, Staged MP, and a lower post-pruning fine-tuning learning rate, Staged MP recovers accuracy and becomes the sparse baseline closest in behavior to the dense reference among the LT baseline, RigL, one-shot MP, and Staged MP. Even there, $\DBEacc$ remains $1.577$ for Staged MP and $2.021$ for RigL, so accuracy recovery still does not imply behavioral interchangeability.

BP-LT selection reduces distance in the broader selection-eligible pool, with selected sources spanning several sparse methods. Together, these rows support two distinct conclusions: Behavior-aware sparse-candidate screening can identify a closer replacement within an evaluated pool, and LT baselines remain behaviorally distinguishable from dense references after accuracy recovery in the measured replacement regimes.

\subsubsection{Interpreting the Recovery Band}
The ResNet-34 result illustrates why accuracy gap and behavioral distance should be read together. Candidates with global random sparse masks have a low behavioral-compatibility distance in this setting and still satisfy the prespecified test recovery band. The accuracy of those candidates, however, is nearly 1~pp below the dense reference and lower than the accuracy of the LT baseline or one-shot MP alternatives. This lower-distance but lower-accuracy pattern reflects our construction: $\DBEacc$ measures differences across behavioral coordinates, while clean aggregate accuracy is evaluated as a separate matching constraint. The explicit accuracy gap and band status therefore remain part of the replacement audit rather than footnotes to the behavioral distance.

\subsubsection{Pretrained and Transfer Settings}
The Imagenette protocols with 224-pixel inputs add evidence from pretrained architectures. With pretrained ConvNeXt-Tiny, the dense reference is highly accurate, and the LT baseline at 50\% sparsity exceeds the reference's clean accuracy, yet dense-threshold decisions, OOD behavior, $\mathrm{PredDisagree}$, and aggregate behavioral distance still move. The pretrained ViT-Tiny protocol at 50\% sparsity adds a transformer-backed case matched under the study-defined accuracy rule: The LT baseline recovers to within about 0.3~pp of the dense reference and still has an $\mathrm{AcceptFlip}_{\tau_D}$ rate of about 8\%.

The fine-grained transfer protocols extend the same pattern beyond the Imagenette protocols. The LT baseline for Flowers-102 at 50\% sparsity recovers to within about 0.3~pp of the dense reference's accuracy while still showing $\DBEacc=0.739$ and an $\mathrm{AcceptFlip}_{\tau_D}$ rate of about $7.3\%$. FGVC-Aircraft at 50\% sparsity similarly recovers within about 1~pp and still shows a behavioral gap that behavior-aware selection reduces. These outcomes extend the behavioral-gap evidence from CIFAR and Imagenette to fine-grained transfer.

\subsection{Related Work}

The lottery ticket hypothesis PASS framework (LTH-PASS) established a beyond-accuracy LT evaluation across shift, uncertainty, interpretability, and loss geometry \citep{chen2022wineverything}. Compression work also preserves per-example decisions or measures answer flips and output distances relative to a baseline model \citep{chee2022model,dutta2024accuracy}. Building on these foundations, our distinction is the inherited operating point: We condition on clean-accuracy recovery, compare each challenger with the corresponding incumbent, and audit the binary routing action produced by applying the same fixed dense-derived threshold to both models. This view connects sparse replacement to backward-compatible model updates. Prior work studies prediction churn, negative flips, downstream pipeline failures, and compatibility-aware objectives or selection \citep{milanifard2016launch,bansal2019updates,srivastava2020backward, yan2021positive,otles2023updating}. Our policy coordinate applies the paired comparison principle to accept--review routing rather than only predicted labels or correctness.

\paragraph{LT recovery.}
The LT hypothesis showed that dense networks can contain sparse subnetworks that, when initialized at their original weights and trained in isolation, recover test accuracy comparable to that of their dense counterparts \citep{frankle2019lottery}. Subsequent work clarified the roles of rewinding, late-training dynamics, and pruning schedules \citep{frankle2020stabilizing,frankle2020linear,renda2020comparing}, while large-scale studies showed that sparse recovery depends strongly on architecture, optimizer, and training recipe \citep{gale2019state}. Work on LT transfer further showed that sparse initializations can generalize across related datasets and optimizers \citep{morcos2019one}. Beyond-accuracy LT evaluation extends this line of work beyond clean-accuracy recovery. Most directly, LTH-PASS asks whether sparse subnetworks can replace dense counterparts after accuracy recovery and evaluates distribution shift, uncertainty, interpretability, and loss geometry \citep{chen2022wineverything}. We build on this foundation by testing compatibility with an inherited incumbent operating point and distinguishing symmetric continuity from signed service-level preservation.

\paragraph{Pruning and sparse training baselines.}
LT methods build on classical and modern pruning methods that remove weights according to saliency or magnitude and retrain the surviving network \citep{lecun1989optimal,han2015learning,han2016deep,zhu2018prune}. Broader surveys and controlled retraining studies caution that compression claims are recipe-dependent \citep{liu2019rethinking,blalock2020state}. Generalization analyses likewise show that pruning effects cannot be reduced to parameter count alone \citep{jin2022pruning}. Pruning at initialization and dynamic sparse training provide competitive sparse alternatives, including SNIP, GraSP, SynFlow, and RigL \citep{lee2019snip,wang2020picking,tanaka2020pruning,evci2020rigging}. Audits of pruning at initialization further show that layerwise sparsity allocation explains much of the advantage over random pruning \citep{frankle2021pruning}. We therefore evaluate LT baselines together with controls based on conventional pruning, pruning at initialization, dynamic sparse training, and random sparsity rather than treating any single sparse construction as definitive.

\paragraph{Reliability after compression.}
Compression can change which examples a network forgets, which groups suffer, and how robust the model is under shift \citep{hooker2020compressed,liebenwein2021lost,tran2022disparate}. Related sparse-reliability work studies calibrated or OOD-robust LTs and sparse training \citep{venkatesh2020calibrate,diffenderfer2021winning,lei2023calibrating}. Privacy-risk diagnostics based on an MIA provide another clean-accuracy-excluded behavioral coordinate for comparing trained models \citep{shokri2017membership,yeom2018privacy,carlini2022membership}. Taken together, prior work already shows that compression can alter example-level outputs and reliability even when aggregate accuracy is similar, with effects that may be adverse or beneficial depending on the diagnostic and recipe \citep{hooker2020compressed,liebenwein2021lost,diffenderfer2021winning,tran2022disparate,chen2022wineverything,chee2022model,dutta2024accuracy,tong2026compression}. Post-hoc pruning benchmarks additionally expose calibration--corruption trade-offs whose directions can depend on architecture and pruning recipe \citep{mitra2024investigating}. The closest compression precedents ask for broad capability parity, preserve per-example decisions during compression, or measure answer flips and output distances relative to a baseline \citep{chen2022wineverything,chee2022model,dutta2024accuracy}. Our contribution centers on the downstream object held fixed: We condition on clean-accuracy recovery, measure paired deviation from a matched dense reference, and test the binary routing action induced by reusing the same dense-derived proxy threshold. This design yields an operating-point compatibility audit and a validation-time screening rule for strict drop-in replacement.

\paragraph{Calibration, OOD, and selective policies.}
Calibration and uncertainty studies show that predictive confidence can be misleading even when in-distribution accuracy is high \citep{naeini2015obtaining,guo2017calibration,ovadia2019trust}. OOD detection methods and common-corruption benchmarks provide operational probes of distribution-shift behavior \citep{hendrycks2017baseline,hendrycks2019benchmarking}. Selective classification studies abstention and review policies rather than forced prediction \citep{geifman2017selective}. These literatures motivate our coordinates for calibration, OOD behavior, corruption accuracy, and dense-threshold policies, which test preservation of the dense reference's operating point after sparse replacement.

Related model-update work measures prediction churn as paired, example-level label disagreement across model versions, including settings in which aggregate accuracy remains similar \citep{milanifard2016launch,bahri2021churn}. Backward-compatibility work further studies new errors, downstream pipeline failures, human expectations, threshold-dependent compatibility, and compatibility-aware objectives or selection \citep{bansal2019updates,srivastava2020backward,yan2021positive,otles2023updating}. Our $\mathrm{AcceptFlip}_{\tau_D}$ coordinate uses the same paired-comparison principle for a different object: the binary accept--review action induced by applying the fixed dense threshold to both models. The acceptance-flip coordinate symmetrically pools both flip directions under the fixed threshold and does not require test labels; unlike prediction churn or a negative flip, the policy coordinate measures preservation of a proxy routing policy rather than a change in predicted class or correctness.

\paragraph{Representations and multiplicity.}
Similarity methods such as CKA show that models with comparable accuracy need not compute the same internal features \citep{kornblith2019similarity}. More broadly, the Rashomon effect view, predictive multiplicity, and underspecification show that high-performing models can make different individual decisions or encode different solutions \citep{breiman2001statistical,marx2020predictive,damour2022underspecification}. We use representation distance and prediction disagreement as behavioral probes alongside calibration, OOD response, dense-threshold policy decisions, privacy-risk diagnostics, and corruption accuracy. Representation distance is a compatibility diagnostic when embeddings or intermediate features are reused by downstream heads, retrieval, or monitoring; the representation diagnostic is not a necessary component of a label-only replacement contract.

\subsection{Limitations and Future Extensions}

\paragraph{Scope and attribution.}
We study unstructured sparse replacements in controlled vision tasks. Extending the audit to ImageNet-scale and language models, structured sparsity, and end-to-end latency and energy remains future work; active parameter counts alone do not establish hardware speedups. We use the incumbent as the sole dense reference, with no independently retrained dense challenger. The results therefore show that clean-accuracy recovery does not certify continuity for the evaluated sparse models, but do not isolate excess incompatibility caused by sparsity from ordinary retraining churn \citep{milanifard2016launch,bahri2021churn,bansal2019updates}. Dense-to-dense controls are needed for that causal comparison.

\paragraph{Audit, selection, and theory.}
$\DBEacc$ summarizes only the active, protocol-specific coordinates and is neither exhaustive nor a signed utility measure. $\mathrm{AcceptFlip}_{\tau_D}$ audits one dense-derived threshold and pools both routing directions. A production audit should predeclare required coordinates, margins, missingness rules, and uncertainty or non-inferiority checks, adding threshold sweeps or directional costs when relevant. BP-LT selects from a finite measured pool and neither guarantees nor constructs a jointly compatible candidate. Broader pools, reference- or threshold-aware training, and constrained or Pareto selection are natural extensions. The theory establishes non-identifiability and a local boundary mechanism; linking its coefficient to finite trained models remains open.

\end{document}